\documentclass{article} 
\usepackage{nips15submit_e,times}

\usepackage{url}

\usepackage{times}
\usepackage{epsfig}
\usepackage{graphicx}
\usepackage{amsmath}
\usepackage{amssymb}
\usepackage{bbold}
\usepackage{dsfont}

\usepackage{enumerate}

\usepackage[toc,page]{appendix}
\usepackage[font=small,labelfont=bf]{caption}

\usepackage{xcolor}
\usepackage{amsthm}
\usepackage{amsfonts}
\usepackage{caption}
\usepackage{subcaption}
\usepackage{bm}
\usepackage{isomath}
\usepackage[numbers]{natbib}
\usepackage{footmisc}
\usepackage{stmaryrd}
\usepackage{fixltx2e}
\usepackage{dblfloatfix}
\usepackage{pbox}
\usepackage{capt-of}
\usepackage[normalem]{ulem}
\usepackage{multirow}

\usepackage{colortbl}

\makeatletter
\@namedef{ver@everyshi.sty}{}
\makeatother
\usepackage{pgf}

\newcommand{\xpt}{\edef\f@size{\@xpt}\rm}

\nipsfinaltrue

\def\ie{\emph{i.e.}}
\def\etc{\emph{etc}}

\renewcommand\vec[1]{\ensuremath\boldsymbol{#1}}
\renewcommand\cdots{...}

\newcommand{\mY}{\boldsymbol{Y}}

\newcommand{\vy}{\boldsymbol{y}}
\newcommand{\vl}{\boldsymbol{l}}
\newcommand{\mL}{\boldsymbol{L}}

\newcommand{\mX}{\boldsymbol{X}}
\newcommand{\mW}{\boldsymbol{W}}
\newcommand{\vx}{\boldsymbol{x}}

\newcommand{\mbr}[1]{\mathbb{R}^{#1}}

\newcommand{\idx}[1]{\mathcal{I}_{#1}}
\newcommand{\semipd}[1]{\mathcal{S}_{+}^{#1}}
\newcommand{\spd}[1]{\mathcal{S}_{++}^{#1}}

\newcommand{\vphi}{\boldsymbol{\phi}}
\newcommand{\vpsi}{\boldsymbol{\psi}}

\newcommand{\bigoh}{\mathcal{O}}

\newcommand{\fnorm}[1]{\left\|{#1}\right\|_F}

\newcommand{\set}[1]{\left\{#1\right\}}

\DeclareMathOperator*{\argmin}{arg\,min}
\DeclareMathOperator*{\argmax}{arg\,max}

\DeclareMathOperator*{\trace}{Tr}

\DeclareMathOperator*{\diag}{diag}

\DeclareMathOperator*{\avg}{Avg}

\newtheorem{proposition}{Proposition}
\newtheorem{remark}{Remark}

\newcommand{\vw}{\boldsymbol{w}}

\def\eg{\emph{e.g.}}

\newcommand{\mIdent}{\boldsymbol{\mathds{I}}}

\newcommand{\vOnes}{\mathbb{1}}

\newcommand{\mK}{\boldsymbol{K}}

\newcommand{\cov}{\boldsymbol{\Sigma}}

\newcommand{\mM}{\boldsymbol{M}}

\newcommand{\stkout}[1]{{\ifmmode\text{\sout{\ensuremath{#1}}}\else\sout{#1}\fi}}

\newcommand{\mA}{\boldsymbol{A}}
\newcommand{\comment}[1]{}

\title{\Large Zero-Shot Kernel Learning}

\author{Hongguang Zhang\thanks{Both authors contributed equally.\newline\indent\indent$\!\!$This work is published in CVPR'18. Please respect the authors' efforts by not copying/plagiarizing bits and pieces of this work for your own gain. If you find anything inspiring in this work, be kind enough to cite it.}\textsuperscript{$\;\,$,2,1}\qquad Piotr Koniusz\textsuperscript{$*$,1,2}\\
$^1$Data61/CSIRO, $^2$Australian National University\\
firstname.lastname@\{data61.csiro.au\textsuperscript{1}, anu.edu.au\textsuperscript{2}\}
}

\newcommand\keywords[1]{}
\nipsfinalcopy 

\begin{document}

\maketitle

\def\arxiv{arxiv}
\begin{abstract}
In this paper, we address an open problem of zero-shot learning. Its principle is based on learning a mapping that associates feature vectors extracted from {\em\ie}~images and attribute vectors that describe objects and/or scenes of interest.
In turns, this allows classifying unseen object classes and/or scenes by matching feature vectors via mapping to a newly defined attribute vector describing a new class. Due to importance of such a learning task, there exist many methods that learn semantic, probabilistic, linear or piece-wise linear mappings. In contrast, we apply well-established kernel methods to learn a non-linear mapping between the feature and attribute spaces. We propose an easy learning objective inspired by the Linear Discriminant Analysis, Kernel-Target Alignment and Kernel Polarization methods \cite{fisher_fda,kernel_alignment,ploarisation} that promotes incoherence. We evaluate performance of our algorithm on the Polynomial as well as shift-invariant Gaussian and Cauchy kernels. Despite simplicity of our approach, we obtain state-of-the-art results on several zero-shot learning datasets and benchmarks including a recent AWA2 dataset \cite{XianCVPR2017}.
\end{abstract}


\section{Introduction}
\label{sec:intro}
The goal of zero-shot approaches is to learn a mapping that matches any given data point, say an image descriptor, to a predefined set of attributes which describe contents of that data point/image, \etc. 
The hope is that such a mapping will generalize well to previously unseen combinations of attributes therefore facilitating recognition of new classes of objects without the need for retraining the mapping itself on these new objects. The quality of zero-shot learning may depend on many factors \ie, (i) the mapping has to match  visual traits captured by descriptors and attributes well, (ii) some visual traits and attributes describing them have to be shared between the classes of objects used for training and testing--otherwise transfer of knowledge is impossible, (iii) the mapping itself should not overfit to the training set.

The task of zero-shot learning can be considered as a form of transfer learning~\cite{transfer_workshop_1995, transfer_workshop_2016}. Given a new (target) task to learn, the arising question is how to identify the so-called commonality \cite{tommasi_cvpr10,me_domain} between this task and previous (source) tasks, and transfer knowledge from the source tasks to the target one. Thus, one has to address three questions: what to transfer, how, and when~\cite{tommasi_cvpr10}. For zero-shot learning, the visual traits and attributes describing them, which are shared between the training and testing sets, form this commonality. However, the objects in training and testing sets are described by disjoint classes. The role of mapping is to facilitate identification of presence/absence of such attributes and therefore enable the knowledge transfer. From that point of view, zero-shot learning assumes that the commonality is pre-defined (attributes typically are) and can be identified in both training and testing data. Alternatively, one can try to capture the commonality from the training and/or testing data (manually or automatically) beforehand \eg, by discovering so-called {\em word2vec} embeddings \cite{word2vec,akata_word2vec}.

Going back to zero-shot learning terminology, we focus in this paper on the design of mapping a.k.a. compatibility function. Various works addressing zero-shot learning and the design of mapping functions have been proposed \cite{larochelle2008zero,farhadi2009describing,rohrbach2011evaluating,patterson2012sun,socher2013zero,norouzi2013zero,frome2013devise, akata2013label}, to name but a few of approaches evaluated in \cite{{xian2017zero}}. We note the use of two kinds of compatibility functions: linear and non-linear. 

\begin{figure}[t]
\centering
%
\centering\includegraphics[trim=0 0 0 0, clip=true, width=8.4cm]{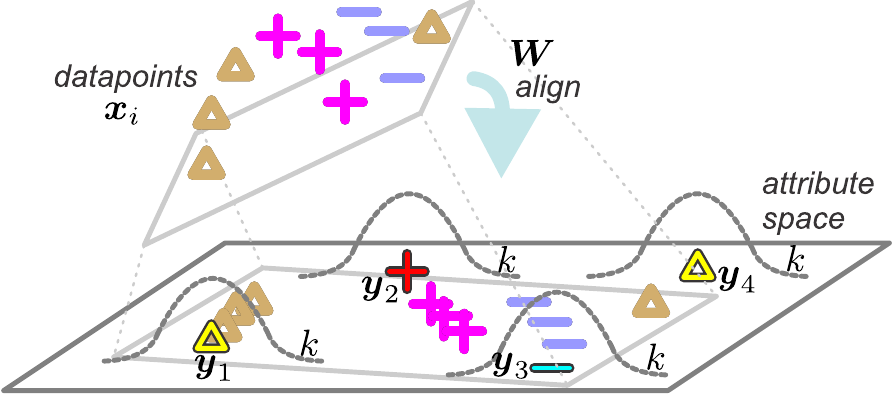}
%
%
\vspace{-0.2cm}
\caption{Zero-shot kernel alignment. Datapoints $\vx$ are projected to the attribute space via the rotation and scaling matrix $\mW$ which we learn. Kernels $k$ are centered at $\vy$ which are attribute vectors.}\vspace{-0.3cm}
\label{fig:principle}
\end{figure}

In this paper, we recognize the gap in the trends and employ kernel methods \cite{scholkopf_rbf} combined with an objective inspired by the Linear Discriminant Analysis (LDA) \cite{fisher_fda,duda_patternclass}, 
a related convex-concave relaxation KISSME \cite{kissme}, Kernel-Target Alignment \cite{kernel_alignment} and Kernel Polarization \cite{ploarisation} methods. Specifically, we are interested in training a mapping function via a non-linear kernel which `elevates' datapoints from the Euclidean space together with attribute vectors into a non-linear high-dimensional Hilbert space in which the classifier can more easily separate datapoints that come from different classes.  
%
Our objective seeks a mapping for which all datapoints sharing the same label with attribute vectors are brought closer in the Hibert space to these attribute vectors while datapoints from different classes are pushed far apart. The mapping function takes a form of projection with soft/implicit weak incoherence mechanism related to idea \cite{ramirez_incoh}.  Thus, our algorithm is somewhat related to subspace selection methods as our projection 
allows rotation and scaling but limits amount of shear and excludes translation. Figure \ref{fig:principle} illustrates our approach. 

For kernels, we experiment with the Polynomial family \cite{scholkopf_kernels} and the shift-invariant Radial Basis Function ({\em RBF}) family including the Gaussian \cite{scholkopf_kernels,scholkopf_rbf} and Cauchy \cite{cauchy_kernel} kernels. Our choice of the Polynomial family is motivated by its simplicity while the RBF family by its ability to embed datapoints in a potentially infinite-dimensional Hilbert space. Moreover, kernels are known to induce implicitly regularization on the plausible set of functions from which a solution to classification problem is inferred \eg, a small radius of an RBF kernel implies a highly complex decision boundary while a large radius has the opposite impact.

To our best knowledge, we are the first to combine non-linear kernels with an objective inspired by LDA \cite{fisher_fda} and kernel alignment \cite{kernel_alignment} in the context of zero-shot learning.

\section{Related Work}
\label{sec:related_work}

\begin{table*}[t]
\vspace{-0.4cm}
\begin{center}
{
\setlength{\tabcolsep}{0.2em}
\centering
\hspace{-0.7cm}
\begin{tabular}{c | c c c c c c c}
\multirow{2}{*}{Kernel} & \multirow{2}{*}{$k(\vx,\vy; \mW)$} & \multirow{2}{*}{$\spd{}$ if} & Shift- &\multirow{2}{*}{$\frac{\partial k(\vx,\vy; \mW)}{\partial\mW}$}\\
& & & \kern-0.4em inv. & &  &  &\\
\hline
Polynomial \cite{scholkopf_kernels} & $\left(\vx^T\mW\vy\!+c\right)^r$ & $\textstyle c\!\geq\!0, r\!\in\!\idx{\infty}$ & no & $r\vx\vy^T\!\left(\vx^T\mW\vy+c\right)^{r-1}$\\
Gaussian \cite{scholkopf_rbf} & $\exp\left(\frac{-||\mW^T\!\vx-\vy||_2^2}{2\sigma^2}\right)$ & $\sigma\!>\!0$ & yes & \kern-0.6em$-\frac{\vx(\mW^T\vx-\vy)^T}{\sigma^2}\exp\left(-\frac{||\mW^T\vx-\vy||_2^2}{2\sigma^2}\right)$\kern-0.9em\\
Cauchy \cite{cauchy_kernel} & $\frac{1}{1+\sigma||\mW^T\!\vx-\vy||_2^2}$ & $\sigma\!>\!0$ & yes & $-\frac{2\sigma\vx(\mW^T\vx\!-\!\vy)^T}{(1+\sigma||\mW^T\vx-\vy||_2^2)^2}$\\
\end{tabular}
}
\end{center}
\caption{Polynomial, Gaussian and Cauchy kernels $k(\vx,\vy; \mW)$, their properties and derivatives w.r.t. the projection matrix $\mW$ introduced by us. Column $\spd{}$ indicates when these kernels are positive definite. Parameters $r$, $c$ and $\sigma$ denote the degree, bias and radius, respectively.}
\label{tab:kernels}
\vspace{-0.3cm}
\end{table*}

We describe first the most popular zero-shot learning methods and explain how our work differs from them.




\noindent{\textbf{Linear mapping}} a.k.a. the linear compatibility function is widely utilized in zero-shot learning. Some notable works include \cite{frome2013devise,akata2013label,akata2015evaluation,romera2015embarrassingly}, to name but a few of methods.

Deep Visual-Semantic Embedding ({\em DeViSE}) model \cite{frome2013devise} utilizes a pre-trained neural language model and a deep neural network later retrained for zero-shot learning. 
DeViSE uses a pairwise ranking loss inspired by Ranking SVM \cite{rank_svm} to learn parameters in the linear transformation layer.

Attribute Label Embedding ({\em ALE}) model \cite{akata2013label} is inspired by the structured prediction approaches. The authors introduce a linear compatibility function and learn its parameters by solving a WSABIE \cite{WSABIE} ranking objective which ensures that more importance is given to the top of the ranking list. 

Structured Joint Embedding ({\em SJE}) framework \cite{akata2015evaluation} employs the linear compatibility function and structured SVM \cite{structured_svm} to give importance only to the top of the ranked list. 
For the latent representation which describes the seen and unseen classes, SJE uses either human annotated attribute vectors or latent {\em word2vec} embeddings \cite{word2vec} or global vectors {\em glove} \cite{glove} learned from a text corpora.

Embarrassingly Simple Zero-Shot Learning ({\em ESZSL}) combines a linear mapping, simple empirical loss and regularization terms which penalize projection of the feature vectors from Euclidean into the attribute space, and projection of the attribute vectors back into the Euclidean space. 

Our approach differs 
in that we do explicitly use a non-linear mapping function. With the use of kernel methods, we first embed datapoints into the attribute space. Then, we utilize a kernel of our choice for scoring the compatibility between a given datapoint and its corresponding attribute. 

\noindent{\textbf{Non-linear methods }} have also been utilized in zero-shot learning \cite{socher2013zero,latem_cvpr16a} and demonstrated improved performance.

Cross Modal Transfer ({\em CMT}) approach \cite{socher2013zero} employs a two-layer neural network with a hyperbolic tangent activating function which maps images into a semantic space of words. 
Unlike many zero-shot learning models, 
CMT works well on a mixture of seen and unseen classes. 

We note that our approach also handles well both seen and unseen classes simultaneously. We also use a non-linear mapping, however, we realize this via a non-linear kernel and subspace learning rather than  neural networks.

Latent Embeddings ({\em LatEm}) model \cite{latem_cvpr16a} 
uses a piece-wise linear compatibility function by combining multiple mappings learned via
a pairwise ranking loss 
similar to the DeViSE model \cite{frome2013devise}. 
In the test time, the scoring function selects a mapping from the learned set which is maximally compatible with a given pair of feature and attribute vectors.

Our approach 
extends easily to learning multiple mapping functions 
such as class-specific subspaces. 
Moreover, our method uses a non-linear kernel that scores how compatible a given pair of feature and attribute vector is.

\noindent{\textbf{Semantic and probabilistic }} approaches to zero-shot learning include Direct and Indirect Attribute Prediction models ({\em DAP}) and ({\em IAP}) \cite{lampert2014attribute} which learn probabilistic attribute classifier and predict a class by combining classifier scores. 
Convex Combination of Semantic Embeddings ({\em ConSE}) \cite{norouzi2013zero} maps images into a so-called semantic embedding space via
convex combination of the class label embedding vectors. Semantic Similarity Embedding ({\em SSE}) \cite{zhang2015zero} represents the source and target data as a mixture of seen class histograms and uses 
a structured scoring function. Synthesized Classifiers ({\em SYNC}) \cite{changpinyo2016synthesized} learn a mapping between a model space and the semantic class embedding space where they use so-called phantom classes. The formulations and larger evaluations of several methods can be found in \cite{XianCVPR2017}. 

\noindent{\textbf{Classifiers. }} 
Linear Discriminant Analysis (LDA) uses two types of statistics: within- and between-class scatters computed from datapoints sharing the same label and the mean vectors of each within-class scatter, respectively. For binary classification, one gets a combined within-class scatter $\cov\!=\!\cov_1\!+\!\cov_2$ and a between-class scatter $\cov^*\!$. Then, LDA seeks a unit length vector $\vw$ to find the direction of maximum and minimum variance of $\cov$ and $\cov^*\!$, respectively: $\argmax_{\vw,||\vw||_2\!=\!1}\vw^T\!(\mM)\vw$ where $\mM\!=\!\cov^{-0.5}\!\cov^*\!\cov^{-0.5}$. 


KISSME \cite{kissme} for metric learning \cite{Kumar_CVPR_2018,Roy_CVPR_2018} assumes  $\cov_S$ for the datapoints considered similar and $\cov_D$ for the dissimilar datapoints. Given two datapoints $\vx$ and $\vx'\!$, 
KISSME is given by $\Delta\vx^T(\mM\!)_{+}\Delta\vx$, where $\Delta\vx\!=\!\vx\!-\!\vx'\!$, $\mM\!=\!\cov_S^{-1}\!-\!\cov_D^{-1}$ and $(\mM\!)_{+}$ is re-projection onto the SPD cone. 

Our objective is related to LDA/KISSME as use within- and between-class terms learnt in a non-linear setting. 

\section{Background}
\label{sec:background}

Below we review our notations and useful kernel tools. 


\subsection{Notations}
\label{sec:notations}
Let $\vx\in\mbr{d}$ be a $d$-dimensional feature vector. $\idx{N}$ stands for the index set $\set{1, 2,\cdots,N}$. 
The Frobenius norm of matrix is given by  $\fnorm{\mX}\!\!=\!\!\!\sqrt{\sum\limits_{m,n} \!\!X_{mn}^2}$, where $X_{mn}$ represents the $\left(m,n\right)$-th element of $\mX$.  
The spaces of symmetric positive semidefinite and definite matrices are $\semipd{d}$ and $\spd{d}$. 
%
Operator $\left[k_{ij}\right]_{i,j\!\in\!\idx{N}}$ denotes stacking coefficients $k_{ij}$ into matrix $\mK$ of size $N\!\times\!N$. 
Moreover, $\delta(x)\!=\!\lim_{\sigma\rightarrow 0}\exp(-x^2/(2\sigma^2))$ returns one for $x\!=\!0$ and zero for $x\!\neq\!0$. We also define $\vOnes\!=\![1,\cdots,1]^T$.

\subsection{Kernel Alignment}
\label{sec:alignment}
Our model relies on the Kernel Alignment and Kernel Polarization methods \cite{kernel_alignment,ploarisation} detailed below. 

\begin{proposition}\label{prop:ka}
Let $k:\mbr{d}\times \mbr{d}\to\mathbb{R}$ and $k':\mbr{d'}\times \mbr{d'}\to\mathbb{R}$ be two positive (semi-)definite kernel functions. Let two data matrices $\mX\!\in\!\mbr{d\times N\!}$ and $\mX'\!\!\in\!\mbr{d'\times N\!}$ contain column vectors $\vx_i\!\in\!\mbr{d}$ and $\vx'_i\!\in\!\mbr{d'}$ for $i\!\in\!\idx{N}$.
Assume two kernel matrices $\mK,\mK'\!\!\in\!\spd{N\times N}\!$ (or $\semipd{}$) for which their $(i,j)$-th element is given by $k(\vx_i,\vx_j)$ and $k'(\vx'_i,\vx'_j)$, respectively, and $i,j\!\in\!\idx{N}$. 
Then, the empirical alignment of two kernels, which also forms a positive (semi-)definite kernel, is the quantity given by the dot-product between $\mK$ and $\mK'\!$:
%
%
\begin{align}
& \left<\mK, \mK'\right>\!=\!\!\!\!\sum_{i,j\in\idx{N}}\!\!k(\vx_i,\vx_j)k'(\vx'_i,\vx'_j).\label{eq:alignment1}
\end{align}
\end{proposition}
\begin{proof}
Mercer theorem \cite{scholkopf_kernels} states that for any $c_i, c_j\!\in\!\mbr{}$, the ineq. $\sum_{i,j\in\idx{N}}c_i c_jk(\vx_i,\vx_j)k'(\vx'_i,\vx'_j)\geq 0$ must hold for a positive (semi-)definite kernel $kk'\!$:
\begin{align}
&\!\!\!\!\textstyle\sum_{i,j\in\idx{N}}\!c_i c_jk(\vx_i,\vx_j)k'(\vx'_i,\vx'_j)=\nonumber\\
&\!\!\!\!\textstyle\sum_{i,j\in\idx{N}}\!c_i c_j\left<\vphi(\vx_i),\vphi(\vx_j)\right>\left<\vpsi(\vx'_i),\vpsi(\vx'_j)\right>=\nonumber\\
&\!\!\!\!\textstyle\left<\sum_{i\in\idx{N}}\!c_i\vphi(\vx_i)\vpsi(\vx'_i)^T,\sum_{j\in\idx{N}}\!c_j\vphi(\vx_j)\vpsi(\vx'_j)^T\right>=\nonumber\\
&\!\!\!\!\textstyle||\sum_{i\in\idx{N}}\!c_i\vphi(\vx_i)\vpsi(\vx'_i)^T||^2_\mathcal{H}\geq 0,
\label{eq:alignment2}
\end{align}
where $\vphi(\vx)\!\in\!\mbr{\hat{d}}$ and $\vpsi(\vx')\!\in\!\mbr{\hat{d}'}\!$ are so-called feature maps \cite{scholkopf_kernels} for kernels $k$ and $k'$. Such maps always exist for positive (semi-)definite kernels by definition \cite{scholkopf_kernels}.
\end{proof}
\begin{remark}
The empirical alignment $\left<\mK,\mK'\right>$ can also be evaluated between rectangular matrices $\mK,\mK'\!\!\in\!\mbr{M\times N}$.
\end{remark}

\begin{proposition}
\label{prop:pol}
Let $k:\mbr{d}\times \mbr{d}\to\mathbb{R}$ be a kernel function which is positive (semi-)definite. Assume that data and kernel matrices $\mX\!\in\!\mbr{d\times N\!}$ and $\mK\!\in\!\spd{N\times N}$ (or $\semipd{}$) are formed as in Prop. \ref{eq:alignment1}. Moreover, let each column vector $\vx_i$ of $\mX$ have a corresponding label $l_i$ for $i\!\in\!\idx{N}$, so that $\mX$ has a corresponding label vector $\vl\!\in\!\{-1,1\}^{N}$. By construction, $\mL\!=
\!\vl\vl^T\!\in\!\semipd{N\times N}$ is a rank-$1$ kernel.
Then, the empirical alignment of kernels $\mX$ and $\mL$, which also forms a positive semidefinite kernel, forms so-called kernel polarization:
\begin{align}
&\!\!\!\!\!\!\!\!\left<\mK, \mL\right>\!=\!\!\!\!\sum_{i,j\in\idx{N}}\!\!l_i\,l_j\,k(\vx_i,\vx_j)\!=\!\vl^T\!\mK\vl
\!=\!\!\!\!\!\!\!\!\sum_{(i,j):\, l_i=l_j}\!\!\!\!\!\!\!k(\vx_i,\vx_j) - \!\!\!\!\!\!\!\!\sum_{(i,j):\, l_i\neq l_j}\!\!\!\!\!\!\!k(\vx_i,\vx_j)  \label{eq:alignment4}
\end{align}
\end{proposition}
\begin{proof}
It follows the same steps as for Proposition \ref{eq:alignment1}. Moreover, $\left<\mK,\mL\right>$ itself forms a positive semidefinite kernel as $\mK$ is positive (semi-)definite and $\mL$ is positive semidefinite (\eg, rank-$1$) by design.
\end{proof}
\begin{proposition}

Assume matrix $\mW\!\in\!\mbr{d\times d}$ which column vectors $\vw_1,\cdots,\vw_{d}\!$ are orthogonal and a projection of datapoint matrix $\mX\!\in\!\mbr{d\times N}$ into the attribute space, that is, $\mY\!=\!\mW^T\!\mX\!\in\!\mbr{d\times N}$. Then the inverse proj. is $\mX\!=\!\mW\mY$.
\label{prop:proj}

\end{proposition}
\begin{proof}
It follows from the orthogonality of column vectors $\vw_1,\cdots,\vw_{d}\!$ of $\mW$ that $\mW^T\!\mW\!=\!\diag(s_1,\cdots,s_{d})$, where $\diag(\vec{s})$ is a diagonal matrix with $s_1,\cdots,s_{d}$ on its diagonal.
\end{proof}

\subsection{Kernel Choices}
\label{sec:kernels}

Table \ref{tab:kernels} details kernels used in this paper, their parameters and properties such as shift-invariance and positive definiteness. We also list derivatives w.r.t. the projection matrix $\mW$ introduced by us to map datapoints to the space of attribute vectors. We detail this in Section \ref{sec:problem}.

\section{Problem Formulation}
\label{sec:problem}

In this section, we detail our zero-shot kernel learning. First, we explain our notations. Let us define data matrices $\mX\!\in\!\mbr{d\times N\!}$ and $\mY\!\in\!\mbr{d'\times N\!}$ which contain $N$ datapoints and attribute vectors as column vectors $\vx_i\!\in\!\mbr{d}$ and $\vy_i\!\in\!\mbr{d'}$ for $i\!\in\!\idx{N}$, respectively. The number of datapoints per class $c\!\in\!\idx{C}$ is $N_c$.  A vector $\vl\!\in\!\idx{C}^{N}$ contains corresponding class labels, one per datapoint/image, \etc.
In our case, datapoints and attribute vectors, which constitute an input to our algorithm, are taken from the standard zero-shot learning package \cite{xian2017zero}. As the available attribute vectors are one per class, we replicate each for all datapoints of a given class. Figure \ref{fig:principle} illustrates our approach. Below we first analyze our proposed weak incoherence mechanism.

\begin{proposition}
\label{prop:inc1}
Let us assume some loss $\ell$ which we minimize w.r.t. projection $\mW$, and $\mX$ and $\mY$ are defined as in Section \ref{sec:problem}. Then, the following expression promotes weak incoherence between column vectors of $\mW$:
\begin{align}\label{eq:weak_incoh}
& \argmin\limits_{\mW} \; \ell\left(||\mW^T\!\mX\!-\!\mY||_2^2\right) + \ell\left( ||\mW\mY\!-\!\mX||_2^2\right).
\end{align}
\end{proposition}
\begin{proof}
For brevity, we drop loss $\ell$ and consider terms $||\mW^T\!\mX\!-\!\mY||_2^2$ and $||\mW\mY\!-\!\mX||_2^2$. Clearly, Eq. \eqref{eq:weak_incoh} will yield approximation error matrices $\Delta\mY$ and $\Delta\mX$ \eg, $\mW^T\!\mX\!=\!\mY\!+\!\Delta\mY$ and $\mW\!\mY\!=\!\mX\!+\!\Delta\mX$. Combining both, we obtain:
\begin{align}\label{eq:weak_incoh2}
& (\mW^T\!\mW\!\!-\!\mIdent)\mY\!=\!\mW^T\!\!\Delta\mX\!\!+\!\Delta\mY\!\!=\!\boldsymbol{\xi},
\end{align}
where $\boldsymbol{\xi}$ is the total approximation error matrix. From Eq. \eqref{eq:weak_incoh2} it follows that if $||\boldsymbol{\xi}||_F^2\!\rightarrow\!0$ then $\mW^T\!\mW\!\rightarrow\!\mIdent$. This means the lower the approximation error is, the higher the incoherence becomes. Moreover, if $||\boldsymbol{\xi}||_F^2\!=\!0$ then columns of $\mW$ are orthogonal w.r.t. each other. Proposition \ref{prop:proj} is a special case of Proposition \ref{prop:inc1}.
\end{proof}

\subsection{Zero-Shot Kernel Alignment}
\label{sec:zero_ker}

For shift-invariant kernels, that is kernels which can be written as $k(\vx\!-\!\vx',0)$, we maximize the following objective which performs kernel alignment for zero-shot learning:
\begin{align}\label{eq:zero_polar}
&\mW^*\!\!=\!\argmax\limits_{\mW}\left<\mK_{\sigma}(\mW)\!+\!\mK'_{\sigma}(\mW), \mL_{\lambda}\right>.\\[-20pt]\nonumber
\end{align}
$\mK_{\!\sigma}(\mW)\!\!\equiv\scriptstyle\left[k_{\sigma}({\mW}^T\!\!\vx_i,\vy_j)\right]_{i,j\!\in\idx{N}}\!\!\!$ and $\mK'_{\!\sigma}(\mW)\!\!\equiv\scriptstyle\left[k_{\sigma}(\vx_i,{\mW}\!\vy_j)\right]_{i,j\!\in\idx{N}}\!\!\!\!$ denote RBF kernels \eg, Gaussian or Cauchy with radius $\sigma$. Note that kernels $\mK_{\sigma}$ and $\mK'_{\sigma}$ use projections ${\mW}^T\!\vx$ and ${\mW}\vy$, respectively, which follow Prop. \ref{prop:inc1}. This implies that Eq. \eqref{eq:zero_polar} is encouraged to find a solution $\mW^*$ for which its column vectors $\vw_1,\cdots,\vw_{d'}\!$ are weakly incoherent/closer to being orthogonal w.r.t. each other. Therefore, our projection matrix $\mW$ is constrained in a soft/implicit manner to be well-regularized. Constrained $\mW$ is closer to being a subspace than an unconstrained $\mW$. As such, we can rotate and scale datapoints to project them into the attribute space. Excluding shear prevents overfitting \eg, it acts  implicitly as a regularization\text{{\color{red}\footnotemark[3]}}. 

We apply polarization detailed in Prop. \ref{prop:pol} which, in some non-linear Hilbert space, will bring closer/push apart all class-related/unrelated datapoints and attribute vectors, respectively. This is also similar in spirit to LDA and KISSME. 
We define $\mL_{\lambda}$ which encodes labels for our polarization inspired zero-shot kernel learning as: %
%
\begin{align}\label{eq:polar_labels}
\mL_{\lambda}\!\equiv\!\left[\delta(l_i\!-\!l_j)\!-\!\lambda(1\!-\!\delta(l_i\!-\!l_j))\right]_{i,j\!\in\idx{N}},
\end{align}
where $l_i$ is the $i$-th coefficient of $\vl$.
By sorting all labels, it can be easily verified that $[\delta(l_i\!-\!l_j)]_{i,j\!\in\idx{N}}$ equals one when $l_i\!=\!l_j$ and that this term contributes equivalent of the block-diagonal entries in $\mL$. Moreover, it moves within-class datapoints close to each other. In contrast, $\lambda\left[(1\!-\!\delta(l_i\!-\!l_j))\right]_{i,j\!\in\idx{N}}$ contributes -$\lambda$ off-diagonally and its role is to move between-class datapoints away from each other. Thus, $\lambda$ controls a form of regularization. It balances positive and negative entries. Depending on labels $\vl$ and $\lambda$, kernel $\mL$ may be positive or negative (semi-)definite\text{{\color{red}\footnotemark[1]}}.

\footnotetext[1]{\label{foot:pd_ex}Compare \eg, $\mL_{0.2}\!=\!\mIdent\!-\!0.2\vOnes\vOnes^T\!\in\!\semipd{5\times5}$ vs. $\mL_{1}\!=\!\mIdent\!-\!\vOnes\vOnes^T\!\in\!\mbr{5\times5}$.}
\footnotetext[2]{\label{foot:idx_choice}If there is one attribute vector per class, the choice of index is fixed.}
\footnotetext[3]{\label{foot:data_mean}We also exclude translation as we mean-center our data/attr. vectors.}

\vspace{0.05cm}
\noindent{\textbf{Complexity.}} For Eq. \eqref{eq:zero_polar}, we obtain $\bigoh(Ndd'\!\!+\!N^2d'\!)$ complexity which reduces to $\bigoh(Ndd'\!\!+\!NCd'\!)$ if one attribute vector per class is defined 
or multiple attribute vectors per class are defined but only one drawn per iteration of SGD as detailed below.

\subsection{Practical Implementation}
\label{sec:ract_case}
Below we show practical expansions of Eq. \eqref{eq:zero_polar} for the RBF kernels from Table \ref{tab:kernels}, which demonstrate the simplicity of our approach: 
\begin{align}\label{eq:zero_polar2}
&\mW^*\!\!=\!\argmax\limits_{\mW}\;\;\left<\mK_{\sigma}(\mW)\!+\!\mK'_{\sigma}(\mW), \mL_{\lambda}\right>=\\
&=\!\!\!\!\!\!\!\!\textstyle\sum\limits_{(i,j):\, l_i=l_j}\!\!\!\!\!\!\!\scriptstyle k_{\sigma}(\mW^T\!\!\vx_i,\vy_j)\!+\!k_{\sigma}(\vx_i,\mW\!\vy_j)
\;\;-\lambda\!\!\!\!\!\!\!\!\!\!\textstyle\sum\limits_{(i,j):\, l_i\neq l_j}\!\!\!\!\!\!\!\scriptstyle k_{\sigma}(\mW^T\!\!\vx_i,\vy_j)\!+\!k_{\sigma}(\vx_i,\mW\!\vy_j)\nonumber\\
&\textstyle=\textstyle\sum\limits_{i\in\idx{N}}\!\!
\scriptstyle N_{l_i}\left(k_{\sigma}(\mW^T\!\!\vx_i,\vy_i)\!+\!k_{\sigma}(\vx_i,\mW\!\vy_i)\right)
\;\;-\lambda\!\!\!\!\!\!\!\!\!\!\textstyle\sum\limits_{(i,j):\, l_i\neq l_j}\!\!\!\!\!\!\!\scriptstyle k_{\sigma}(\mW^T\!\!\vx_i,\vy_j)\!+\!k_{\sigma}(\vx_i,\mW\!\vy_j)\nonumber.
\end{align}
We simplify our problem in Eq. \eqref{eq:zero_polar2} to work with SGD:
\begin{align}\label{eq:zero_polar3}
&f_i(\mW)=
N_{l_i}\!\left(k'_{\sigma}(\mW^T\!\!\vx_i,\vy_i)\!+\!k'_{\sigma}(\vx_i,\mW\!\vy_i)\right)\\
&\qquad\qquad+\lambda\!\!\!\!\!\!\!\!\!\!\textstyle\sum\limits_{j\in\text{Rnd}(\idx{C}\!\setminus\!\{l_i\})}\!\!\!\!\!\!\!k''_{\sigma}(\mW^T\!\!\vx_i,\vy_j)\!+\!k''_{\sigma}(\vx_i,\mW\!\vy_j),\nonumber
\end{align}
where $\mW_{t+1}\!:=\!\mW_{t}\!-\!\frac{\beta_t}{I}\!\sum_{i\in B_t}\!\frac{1}{\sqrt{\mA_t}}\!\frac{\partial f_i(\mW)}{\partial\mW}$ is an SGD-based update for $\mW$, $B_t$ are mini-batches of size $I$, $\beta_t$ is a decaying learning rate, operator $\text{Rnd}(\idx{C}\!\setminus\!\{l_i\})$ selects one index $j\!\in\!\idx{N}$ per class $c\!\in\!\idx{C}\!\setminus\!\{l_i\}$ \eg, if $C\!=\!40$, we get $39$ indexes\text{{\color{red}\footnotemark[2]}} each one randomly sampled per class for all 39 classes. This way, we are able to reduce the complexity as detailed earlier. 
Setting $k'_{\sigma}\!=\!(1\!-\!k_{\sigma})^2$ and $k''_{\sigma}\!=\!k_{\sigma}^2$ 
instead of $k'_{\sigma}\!=\!-k_{\sigma}$ and  $k''_{\sigma}\!=\!k_{\sigma}$, respectively, results in a slightly faster convergence of our algorithm. Also, we set $N_{l_i}\!=\!\frac{N}{C}$ for simplicity. 
Moreover, $\mA_t\!=\!\gamma\mA_{t-1}\!+\!(1\!-\!\gamma)\frac{1}{I}\!\sum_{i\in B_t}\big(\frac{\partial f_i(\mW)}{\partial\mW}\big)^2$ defines so-called moving average of the squared gradient used in the Root Mean Square Propagation (RMSprop) \cite{rmsprop} solver. 

\vspace{0.05cm}
\noindent{\textbf{No incoherence.}} Equations for RBF kernels with no soft/implicit incoherence on $\mW$ can be easily derived from Prop. \ref{prop:pol} by maximizing $\left<\mK_{\sigma}(\mW), \mL_{\lambda}\right>$ w.r.t. $\mW$. This yields a solution similar to Eq. \eqref{eq:zero_polar3}:
\begin{align}\label{eq:zero_polar5}
&f_i(\mW)=
N_{l_i}k'_{\sigma}(\mW^T\!\!\vx_i,\vy_j)\!
+\lambda\!\!\!\!\!\!\!\!\!\!\textstyle\sum\limits_{j\in\text{Rnd}(\idx{C}\!\setminus\!\{l_i\})}\!\!\!\!\!\!\!\!\!\!k''_{\sigma}(\mW^T\!\!\vx_i,\vy_j).
\end{align}

\vspace{0.05cm}
\noindent{\textbf{Polynomial kernel.}} 
%
%
%
As $K\!+\!K'\!=\!2K$ for Polynomial kernels, the objective in Eq. \eqref{eq:zero_polar} cannot implicitly impose weak incoherence constraints on column vectors of $\mW$. Thus, we add a soft penalty on $\mW$ to promote incoherence, adjusted via variable $\alpha$, and we define a modified problem: 
\begin{align}\label{eq:zero_polar_poly}
&\mW^*\!\!=\!\argmax\limits_{\mW}\left<\mK_{\sigma}(\mW), \mL_{\lambda}\right>\!-\!\alpha||\mW^T\!\mW||_F^2\!+\!\alpha\trace(\mW^T\!\mW).
\end{align}
\subsection{Classification}
\label{sec:test_time}
Having learned $\mW^*\!\!$, at the classification stage we apply a simple maximization over testing attribute vectors:
\begin{align}
& \argmax\limits_{j\in\idx{N^*}:\, l^*_j\in\idx{C^*}}k_{\sigma}\left({\mW^*}^T\!\!\vx^*\!,\vy^*_{j}\right)+k_{\sigma}\left(\vx^*\!,{\mW^*}\!\vy^*_{j}\right),
\end{align}
where $k$ is a kernel used during training, $\vx^*\!$ is a testing datapoint. Moreover, $\vy^*\*$ are typically previously unseen testing attribute vectors while $\vl^*$ contains testing labels (typically disjoint with $\vl$). Variable $C^*\!$ is the number of testing classes and $N^*\!$ is the number of testing attribute vectors (typically $N^*\!\!=\!C^*\!$). For the problems which require the Nearest Neighbor classifier with a dot-product based similarity measure, one can apply \eg~the Nystr\" om approximation \cite{scholkopf_kernels} which linearizes kernel $k$ via feature maps $\vphi(\vx)\!\in\!\mbr{\hat{d}}$ and $\vphi'(\vx)\!\in\!\mbr{d}$:
\begin{align}
& k({\mW^*}^T\!\!\!\vx^*\!,\vy^*_{j})\!\approx\!\big<\vphi({\mW^*}^T\!\!\!\vx^*\!),\vphi(\vy^*_{j})\big>+\big<\vphi'(\vx^*\!),\vphi'({\mW^*}\!\vy^*_{j})\big>.
\end{align}
This is however outside of the scope of our evaluations and will be explored in our future work.

\section{Experiments}
\label{sec:expts}

In what follows, we explain our experimental setup followed by evaluations of the proposed zero-shot kernel learning approach. Subsequently, we discuss our findings.


\ifdefined\arxiv
\newcommand{\SrcImgWW}{0.144}
\newcommand{\SrcImgHHH}{2cm}
\newcommand{\SrcImgWWW}{2cm}
\else
\newcommand{\SrcImgWW}{0.240}
\newcommand{\SrcImgHHH}{2cm}
\newcommand{\SrcImgWWW}{2cm}
\fi

\begin{figure}[t]
\centering
\begin{subfigure}[b]{\SrcImgWW\linewidth}
\centering\includegraphics[trim=0 0 0 0, clip=true,width=\SrcImgWWW, height=\SrcImgHHH]{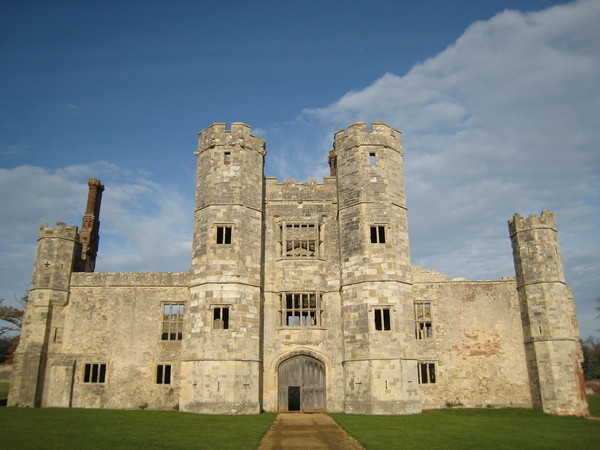}
\end{subfigure}
\begin{subfigure}[b]{\SrcImgWW\linewidth}
\centering\includegraphics[trim=0 0 0 0, clip=true,width=\SrcImgWWW, height=\SrcImgHHH]{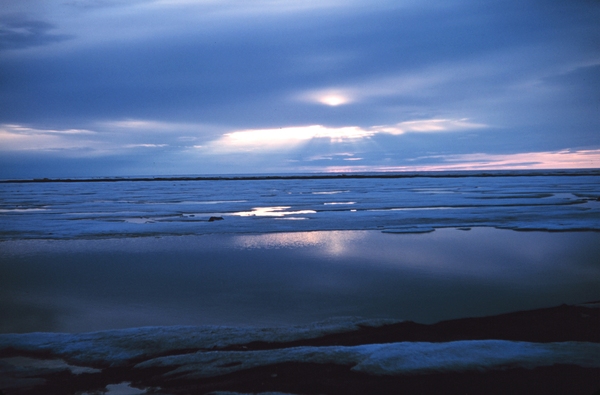}
\end{subfigure}
\begin{subfigure}[b]{\SrcImgWW\linewidth}
\centering\includegraphics[trim=0 0 0 0, clip=true,width=\SrcImgWWW, height=\SrcImgHHH]{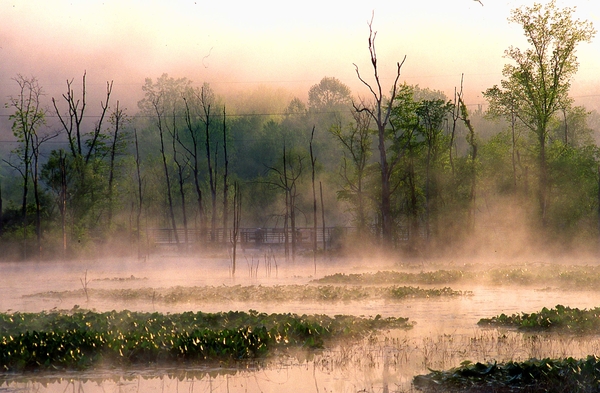}
\end{subfigure}
\begin{subfigure}[b]{\SrcImgWW\linewidth}
\centering\includegraphics[trim=0 0 0 0, clip=true,width=\SrcImgWWW, height=\SrcImgHHH]{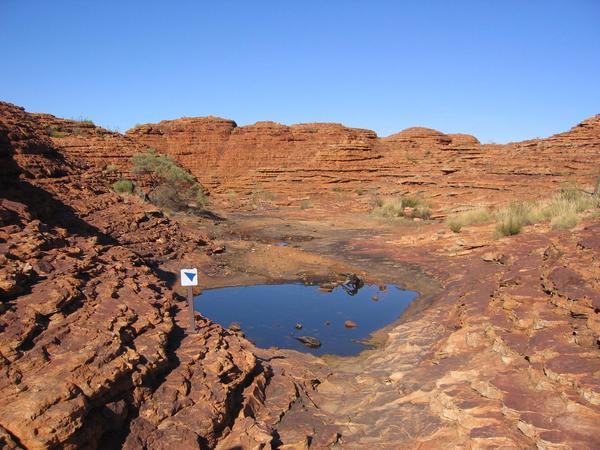}
\end{subfigure}
\\
\vspace{0.052cm}
\begin{subfigure}[b]{\SrcImgWW\linewidth}
\centering\includegraphics[trim=0 0 0 0, clip=true,width=\SrcImgWWW, height=\SrcImgHHH]{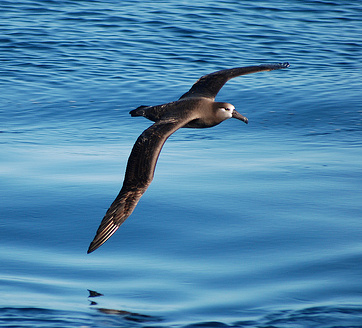}
\end{subfigure}
\begin{subfigure}[b]{\SrcImgWW\linewidth}
\centering\includegraphics[trim=0 0 0 0, clip=true,width=\SrcImgWWW, height=\SrcImgHHH]{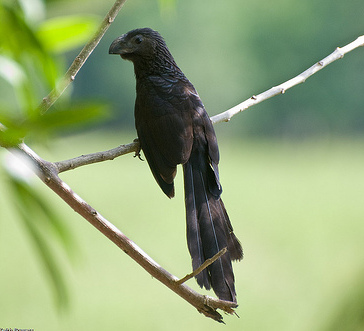}
\end{subfigure}
\begin{subfigure}[b]{\SrcImgWW\linewidth}
\centering\includegraphics[trim=0 0 0 0, clip=true,width=\SrcImgWWW, height=\SrcImgHHH]{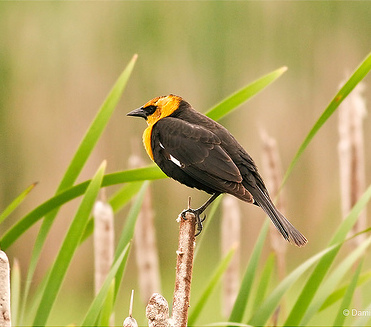}
\end{subfigure}
\begin{subfigure}[b]{\SrcImgWW\linewidth}
\centering\includegraphics[trim=0 0 0 0, clip=true,width=\SrcImgWWW, height=\SrcImgHHH]{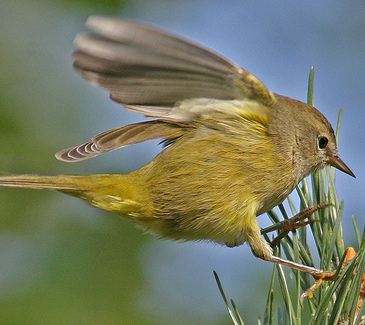}
\end{subfigure}
%
%
\caption{The top and bottom row include example images from the SUN and CUB datasets, respectively. The first two and the last two columns show examples of training and testing images.
}\vspace{-0.25cm}
\label{fig:datasets}
\end{figure}

\vspace{0.05cm}
\noindent{\textbf{Datasets.}} 
%
We use five datasets frequently applied in evaluations of zero-shot learning algorithms. Attribute Pascal and Yahoo ({\em aPY}) \cite{farhadi2009describing} is a small-scale dataset which contains 15339 images, 64 attributes and 32 classes. The 20 classes known from Pascal VOC \cite{pascal07} are used for training and 12 classes collected from Yahoo! \cite{farhadi2009describing} are used for testing. Animals with Attributes ({\em AWA1}) \cite{lampert2014attribute} contains 30475 images of 50 classes. It has a standard zero-shot learning split with 40 training classes and 10 test classes. Each class is annotated with 85 attributes. At present, the original images of AWA1 are not available due to copyrights, therefore, a new version of Animals with Attributes ({\em AWA2}) was proposed in work \cite{XianCVPR2017}. AWA2 also has 40 classes for training and 10 classes for testing. Caltech-UCSD-Birds 200-2011 ({\em CUB}) \cite{welinder2010caltech} has 11788 images, 200 classes and 312 attributes to describe each class. The SUN dataset ({\em SUN}) \cite{patterson2012sun}  consists of 14340 images from 717 classes which are annotated with 102 attributes. Figure \ref{fig:datasets} shows example images from dataset.

We note that a recent evaluation paper \cite{xian2017zero} reports that some of the testing classes from the standard zero-shot learning datasets overlap with the classes of ImageNet \cite{ILSVRC15} which is typically used for fine-tuning image embeddings. This results in a biased evaluation protocol which favors such classes. Therefore, we use newly proposed splits \cite{xian2017zero} for the above five datasets to prevent such a bias and make results of our work follow protocols that offer a fair comparability with the latest state-of-the-art methods.

\vspace{0.05cm}
\noindent{\textbf{Parameters.}} 
In this paper, we make use of the data available in \cite{xian2017zero}. For image embeddings, we use the 2048 dimensional feature vectors extracted from top-layer pooling
units of ResNet-101 \cite{resnet} which was pre-trained on the ImageNet dataset \cite{ILSVRC15}. For class embeddings, we use real-valued per-class attribute vectors provided for the aPY, AWA1, AWA2, CUB and SUN datasets. We perform the mean subtraction and the $\ell_2$-norm normalization on the datapoints and attribute vectors, respectively. 

To learn $\mW\!$, we applied SGD with mini-batches of size $I\!=\!10$, we set the moving average of the squared gradient used in the Root Mean Square Propagation (RMSprop) \cite{rmsprop} solver to $\gamma\!=\!0.99$ and ran the solver for 5--10 epochs. Kernel parameters $\sigma$ and $\lambda$ were selected via cross-validation. For the Polynomial kernel, we chose  $r\!=\!2,4,6$ and parameter $\alpha\!=\!1$. The bias $c$ was selected via cross-validation.

\vspace{0.05cm}
\noindent{\textbf{Testing protocols.}} 
To evaluate our algorithms, we follow two standard protocols as detailed in \cite{xian2017zero}. 
Firstly,  we report the mean top-$1$ accuracy when training on the training data and testing on the classes unseen at the training time. Next, for generalized zero-shot learning protocol, we perform testing on classes both seen and unseen during the training. For that, we use the harmonic mean of training and test accuracies as advocated by \cite{xian2017zero}:
\begin{align}\label{eq:harmonic}
& H = 2\frac{\text{Acc}_{S}\cdot\text{Acc}_{U}}{\text{Acc}_{S}+\text{Acc}_{U}},
\end{align}
where $\text{Acc}_{S}$ and $\text{Acc}_{U}$ denote the accuracy for the seen and unseen at the training stage classes. This strategy can flag up algorithms which overfit to either seen or unseen classes.

\vspace{0.05cm}
\noindent{\textbf{Baselines.}}
We compare our zero-shot kernel learning approach to several works which include: DAP and IAP \cite{lampert2014attribute}, CONSE \cite{norouzi2013zero}, CMT \cite{socher2013zero},  SSE \cite{zhang2015zero},  LATEM \cite{latem_cvpr16a},  ALE \cite{akata2013label}, DEVISE \cite{frome2013devise}, SJE \cite{akata2015evaluation},  ESZSL \cite{romera2015embarrassingly}, SYNC \cite{changpinyo2016synthesized} and SAE \cite{sae}. Section \ref{sec:related_work} contains more details about these methods.

\vspace{0.05cm}
\noindent{\textbf{Our methods.}} We compare the following approaches to the state of the art: the Polynomial kernels denoted ({\em Polynomial}) for $\alpha\!=\!1$ and degree $r\!=\!2,4,6$, respectively, which are combined with Eq. \eqref{eq:zero_polar_poly}. Moreover, we evaluate the RBF kernels such as ({\em Cauchy}) and ({\em Gaussian}) whihc are combined with the formulation in Eq. \eqref{eq:zero_polar5}. Lastly, we evaluate the Cauchy and Gaussian kernels ({\em Cauchy-Ort}) and ({\em Gaussian-Ort}) combined with our soft/implicit incoherence formulation according to Eq. \eqref{eq:zero_polar} and \eqref{eq:zero_polar3}.

\subsection{Evaluations}
\label{sec:eval}

\begin{table}
\setlength{\tabcolsep}{0.3em}
\centering
\begin{tabular}{l l |c c c c c|c}
Method        &                                          &  AWA1 &  AWA2 &  SUN  &  CUB  & aPY & \footnotesize\pbox{3cm}{Better\\than\\SOA} \\ \hline
{\em DAP}     &\kern-0.6em\cite{lampert2014attribute}    & 44.1 & 46.1 & 39.9 & 40.0 & 33.8 & \\ 
{\em IAP}     &\kern-0.6em\cite{lampert2014attribute}    & 35.9 & 35.9 & 19.4 & 24.0 & 36.6 & \\
{\em CONSE}   &\kern-0.6em\cite{norouzi2013zero}         & 45.6 & 44.5 & 38.8 & 34.3 & 26.9 & \\
{\em CMT}     &\kern-0.6em\cite{socher2013zero}          & 39.5 & 37.9 & 39.9 & 34.6 & 28.0 & \\
{\em SSE}     &\kern-0.6em\cite{zhang2015zero}           & 60.1 & 61.0 & 51.5 & 43.9 & 34.0 & \\
{\em LATEM}   &\kern-0.6em\cite{latem_cvpr16a}           & 55.1 & 55.8 & 55.3 & 49.3 & 35.2 & \\
{\em ALE}     &\kern-0.6em\cite{akata2013label}          & 59.9 & 62.5 & 58.1 & 54.9 & 39.7 & \\
{\em DEVISE}  &\kern-0.6em\cite{frome2013devise}         & 54.2 & 59.7 & 56.5 & 52.0 & 39.8 & \\
{\em SJE}     &\kern-0.6em\cite{akata2015evaluation}     & 65.6 & 61.9 & 53.7 & 53.9 & 32.9 & \\
{\em ESZSL}   &\kern-0.6em\cite{romera2015embarrassingly}& 58.2 & 58.6 & 54.5 & 53.9 & 38.3 & \\
{\em SYNC}    &\kern-0.6em\cite{changpinyo2016synthesized}&54.0 & 46.6 & 56.3 & 55.6 & 23.9 & \\
{\em SAE}     &\kern-0.6em\cite{sae}                     & 53.0 & 54.1 & 40.3 & 33.3 &  8.3 & \\
\hline
{\em Polynomial}, $r\!=\!2$\kern-1.7em & & 66.2 & 64.9 & 58.7 & 54.5 & 41.6 & \textbf{4/5}\\
{\em Polynomial}, $r\!=\!4$\kern-1.7em & & 65.0 & 64.3 & 59.7 & \textbf{57.1} & 41.7 & \textbf{4/5}\\
{\em Polynomial}, $r\!=\!6$\kern-1.7em & & 63.6 & 63.3 & 59.2 & 54.6 & 41.5 & 3/5\\
{\em Cauchy}                          & & 60.1 & 58.3 & 57.8 & 48.7 & 35.2 & 0/5\\
{\em Cauchy-Ort}                      & & \textbf{71.0} & 69.9 & 60.4 & 49.3 & 41.9 & \textbf{4/5}\\
{\em Gaussian}                        & & 60.5 & 61.6 & 60.6 & 52.2 & 38.9 & 1/5\\
{\em Gaussian-Ort}                    & & 70.1 & \textbf{70.5} & \textbf{61.7} & 51.7 & \textbf{45.3} & \textbf{4/5}\\
\hline
\end{tabular}
\caption{Evaluations on the standard protocol and the newly proposed datasplits. ({\em Better than SOA}) column indicates the number of datasets on which our methods outperform the state-of-the-art methods listed in the upper part of the table.}
\label{tab:tab1}\ifdefined\arxiv\else\vspace{-0.3cm}\fi
\end{table}

\begin{table*}[t]
\setlength{\tabcolsep}{0.4em}
\ifdefined\arxiv\fontsize{7.25}{9}\selectfont\else\fi
\centering
%
\begin{tabular}{ll|ccc|ccc|ccc|ccc|ccc|c}
                                                        & & \multicolumn{3}{c}{{AWA1}} & \multicolumn{3}{c}{{AWA2}} & \multicolumn{3}{c}{{SUN}}  & \multicolumn{3}{c}{{CUB}}  & \multicolumn{3}{c|}{{aPY}} & \multirow{2}{*}{\pbox{3cm}{\footnotesize \vspace{-0.1cm}Better\\[-3pt] than\\[-3pt] SOA}} \\
Method                                                  & & {\em ts} & {\em tr} & {\em H} & {\em ts} & {\em tr} & {\em H} & {\em ts} & {\em tr} & {\em H} & {\em ts} & {\em tr} & {\em H} & {\em ts} & {\em tr} & {\em H} & \\
\hline
{\em DAP}     &\kern-0.6em\cite{lampert2014attribute}     & 0.0&88.7&0.0 & 0.0&84.7&0.0 & 4.2&25.1&7.2 & 1.7&67.9&3.3 & 4.8&78.3&8.0 \\ 
{\em IAP}     &\kern-0.6em\cite{lampert2014attribute}     & 2.1&78.2&4.1 & 0.9&87.6&1.8 & 1.0&37.8&1.8 & 0.2&72.8&0.4 & 5.7&65.6&10.4\\
{\em CONSE}   &\kern-0.6em\cite{norouzi2013zero}          & 0.4&88.6&0.8 & 0.5&90.6&1.0 & 6.8&39.9&11.6 & 1.6&72.2&3.1 & 0.0&91.2&0.0 \\ 
{\em CMT}     &\kern-0.6em\cite{socher2013zero}           & 0.9&87.6&1.8 & 0.5&90.0&1.0 & 8.1&21.8&11.8 & 7.2&60.1&8.7 & 1.4&85.2&2.8 \\ 
{\em CMT*}     &\kern-0.6em\cite{socher2013zero}          & 8.4&86.9&15.3 & 8.7&89.0&15.9& 8.7&28.0&13.3 & 4.7&60.1&8.7 & 10.9&74.2&19.0 \\
{\em SSE}     &\kern-0.6em\cite{zhang2015zero}            & 7.0&80.5&12.9 & 8.1&82.5&14.8 & 2.1&36.4&4.0 & 8.5&46.9&14.4 & 0.2&78.9&0.4 \\ 
{\em LATEM}   &\kern-0.6em\cite{latem_cvpr16a}            & 7.3&71.7&13.3 & 11.5&77.3&20.0 & 14.7&28.8&19.5 & 15.2&57.3&24.0 & 0.1&73.0&0.2\\
{\em ALE}     &\kern-0.6em\cite{akata2013label}           & 16.8&76.1&27.5 & 14.0&81.8&23.9 & 21.8&33.1&26.3 & 23.7&62.8&34.4 &  4.6&73.7&8.7 \\ 
{\em DEVISE}  &\kern-0.6em\cite{frome2013devise}          & 13.4&68.7&22.4 & 17.1&74.7&27.8 & 16.9&27.4&20.9 & 23.8&53.0&32.8 & 4.9&76.9&9.2\\ 
{\em SJE}     &\kern-0.6em\cite{akata2015evaluation}      & 11.3&74.6&19.6 & 8.0&73.9&14.4 & 14.7&30.5&19.8 & 23.5&59.2&33.6 & 3.7&55.7&6.9\\ 
{\em ESZSL}   &\kern-0.6em\cite{romera2015embarrassingly} & 6.6&75.6&12.1 & 5.9&77.8&11.0 & 11.0&27.9&15.8 & 12.6&63.8&21.0 & 2.4&70.1&4.6\\ 
{\em SYNC}    &\kern-0.6em\cite{changpinyo2016synthesized}& 8.9&87.3&16.2 & 10.0&90.5&18.0 & 7.9&43.3&13.4 & 11.5&70.9&19.8 & 7.4&66.3&13.3\\ 
{\em SAE}     &\kern-0.6em\cite{sae}                      & 1.8&77.1&3.5 & 1.1&82.2& 2.2& 8.8&18.0&11.8 & 7.8&54.0&13.6 & 0.4&80.9&0.9 \\
\hline
{\em Polynomial}, $r\!=\!2$\kern-0.6em                   & & 5.8&77.3&10.7 & 6.4&78.8&11.8 & 20.6&31.5&24.9  & 16.7&61.3&26.2 & 4.8&77.5&9.0  & 0/5\\
{\em Polynomial}, $r\!=\!4$\kern-0.6em                   & & 5.7&78.7&10.6 & 7.0&83.0&13.0 & 20.0&31.7&24.5 & \textbf{24.2}&63.9&\textbf{35.1} & 5.7&79.2&10.6  & 1/5\\
{\em Polynomial}, $r\!=\!6$\kern-0.6em                   & & 8.3&78.1&15.0 & 8.7&81.6&15.7 & 21.0&31.0&25.1 & 23.8&58.6&33.8 & 4.9&78.3&9.2  & 0/5 \\
{\em Cauchy}                                            & & 6.0&79.9&11.1 & 6.2&82.7&11.5 & 16.1&29.7&20.9 & 18.2&49.6&26.6 & 1.0&84.9&2.0  & 0/5 \\
{\em Cauchy-Ort}                                        & & \textbf{18.3}&79.3&\textbf{29.8} & 17.6&80.9&29.0 & 19.8&29.1&23.6  & 19.9&52.5&28.9 & 11.9&76.3&\textbf{20.5} & \textbf{3/5}  \\
{\em Gaussian}                                          & & 6.1&81.3&11.4 & 7.3&79.1&13.3 & 18.2&33.2&23.5 & 17.5&59.9&27.1 & 3.0&82.3&5.8 & 0/5 \\
{\em Gaussian-Ort}                                      & & 17.9&82.2&29.4 & \textbf{18.9}&82.7&\textbf{30.8} & 20.1&31.4&24.5 & 21.6&52.8&30.6 &10.5&76.2&18.5 & 2/5 \\ \hline
\end{tabular}
\caption{Evaluations on the generalized zero-shot learning protocol and the newly proposed datasplits. We indicate the mean top-$1$ accuracy on ({\em tr}) train+test classes and ({\em ts}) test classes only. Moreover, ({\em Better than SOA}) indicates the number of datasets on which our methods outperform the other state-of-the-art methods (the upper part of the table) according to the harmonized score ({\em H}). }
\label{tab:tab2}\vspace{-0.1cm}
\end{table*}

We start our evaluations on the standard protocol followed by the generalized zero-shot learning protocol. Subsequently, we perform a sensitivity analysis of our model w.r.t. its hyperparameters given the validation and testing data to demonstrate robustness of zero-shot kernel learning. The results presented in Tables \ref{tab:tab1} and \ref{tab:tab2} were obtained via cross-validation to prevent overfitting to the testing data.

\vspace{0.05cm}
\noindent{\textbf{Standard protocol.}} Table \ref{tab:tab1} lists our results and indicates the scores attained by other recent approaches. Firstly, we note that the Polynomial kernels ({\em Polynomial}, $r\!=\!2$) and ({\em Polynomial}, $r\!=\!4$), and the RBF kernels ({\em Cauchy-Ort}) and ({\em Gaussian-Ort}) for which we imposed soft/implicit incoherence constraints on $\mW\!$, outperform the state-of-the-art approaches (the upper part of the table) on 4 out of 5 datasets (indicated by 4/5 in the table). In contrast, kernels ({\em Cauchy}) and ({\em Gaussian}), for which we imposed no such constraints, perform notably worse. 
This validates the benefits of decoherence in our model. Moreover, we note that the Gaussian kernel ({\em Gaussian-Ort}) attains the best results on 3 out of 5 datasets (highlighted in bold in the table) when compared to our ({\em Cauchy-Ort}) and ({\em Polynomial}, $r\!=\!4$). 
Result-wise,  ({\em Gaussian-Ort}) outperforms the other state-of-the-art methods on AWA1, AWA2, SUN and aPY by 4.5, 8, 3.6, 5.5\% top-$1$ accuracy. In contrast, ({\em Polynomial}, $r\!=\!4$) outperforms other state-of-the-art approaches on CUB by 2.2\%. 

We conjecture that the the good performance of the Gaussian kernel can be attributed to its ability to `elevate' datapoints to a potentially infinite-dimensional Hilbert space where a decision boundary separating datapoints according to labels can be found with ease \eg, linearly non-separable datapoints become separable. We expect that the Cauchy kernel may also be beneficial due to its slowly decaying tails (compared to Gaussian) which result in stronger non-local influences. Lastly, we expect that non-linearity of Polynomial kernels of degree $r\!=\!4$  may also be sufficient for separating otherwise linearly non-separable data.

\vspace{0.05cm}
\noindent{\textbf{Generalized protocol.}} Table \ref{tab:tab2} presents our results on the generalized zero-shot learning protocol. Firstly, we note that our ({\em Cauchy-Ort}) approach outperforms other state-of-the-art approaches on 3 out of 5 datasets (indicated by 3/5 in the table) according to the generalized score ({\em H}) which takes into account the quality of zero-shot learning when testing it on classes which were seen and unseen during the training. Our ({\em Cauchy-Ort}) is closely followed in terms of scores by ({\em Gaussian-Ort}) and ({\em Polynomial}, $r\!=\!4$) which outperform other state-of-the-art methods on 2 and 1 out of 5 datasets, respectively. Moreover, ({\em Cauchy-Ort}) attains the best results on AWA1 and aPY (highlighted in bold in the table) when compared to our ({\em Gaussian-Ort}) which performs the best on AWA2. Number-wise,  ({\em Gaussian-Ort}) outperforms the other state-of-the-art methods (the upper part of the table) on AWA1, AWA2 and aPY by 2.3, 1.2 and 2.5\% measured according to the generalized score ({\em H}).

\ifdefined\arxiv
\newcommand{\SensWW}{0.246}
\newcommand{\SensHHH}{2.8cm}
\newcommand{\SensHHHH}{2.74cm}
\newcommand{\SensWWW}{2cm}
\else
\newcommand{\SensWW}{0.246}
\newcommand{\SensHHH}{3.7cm}
\newcommand{\SensHHHH}{3.55cm}
\newcommand{\SensWWW}{2cm}
\fi

\begin{figure*}[t]
\centering
\hspace{-0.3cm}
\begin{subfigure}[b]{\SensWW\linewidth}
\centering\includegraphics[trim=0 0 0 0, clip=true,height=\SensHHH]{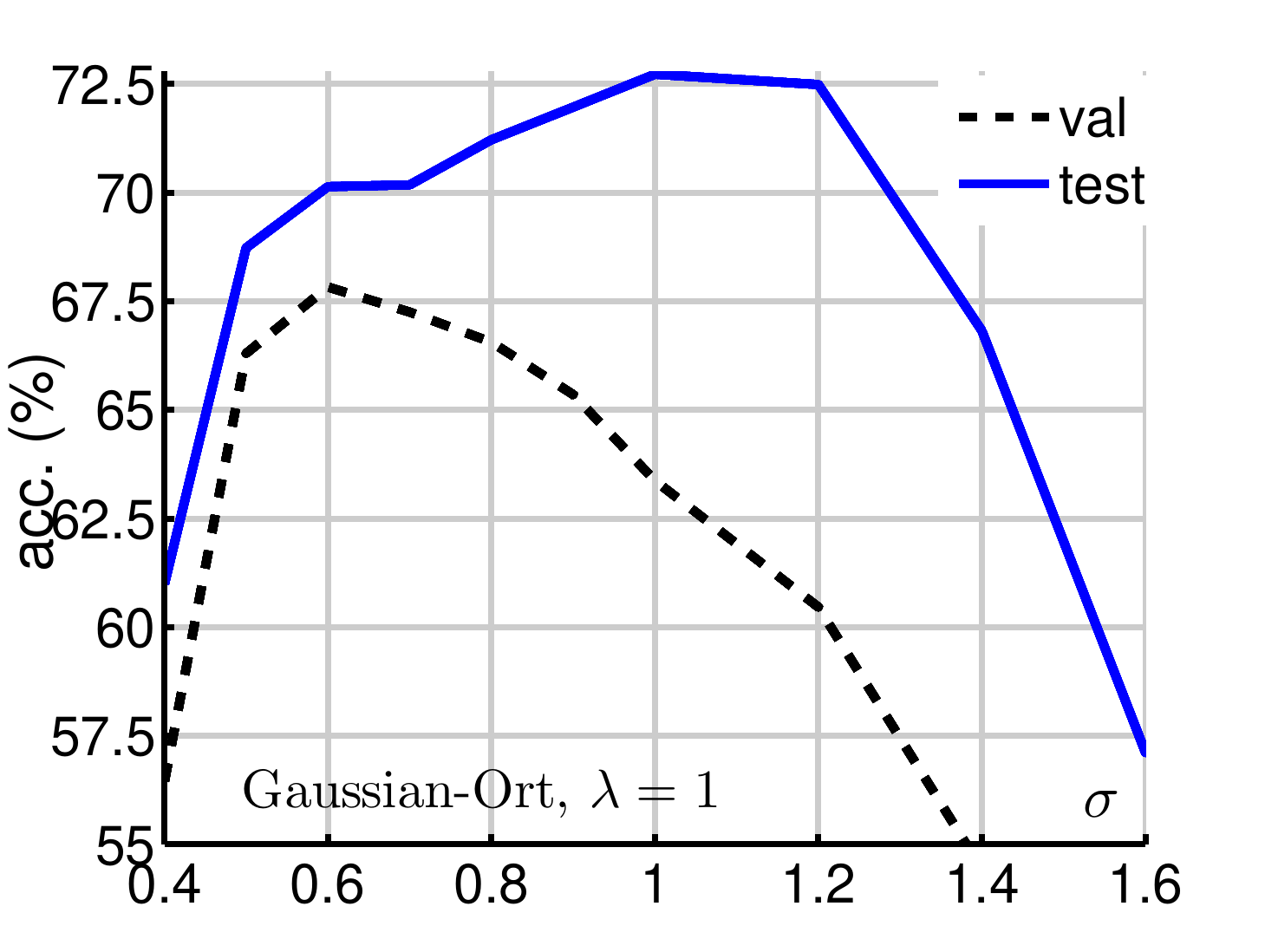}
\end{subfigure}
\begin{subfigure}[b]{\SensWW\linewidth}
\centering\includegraphics[trim=0 0 0 0, clip=true,height=\SensHHH]{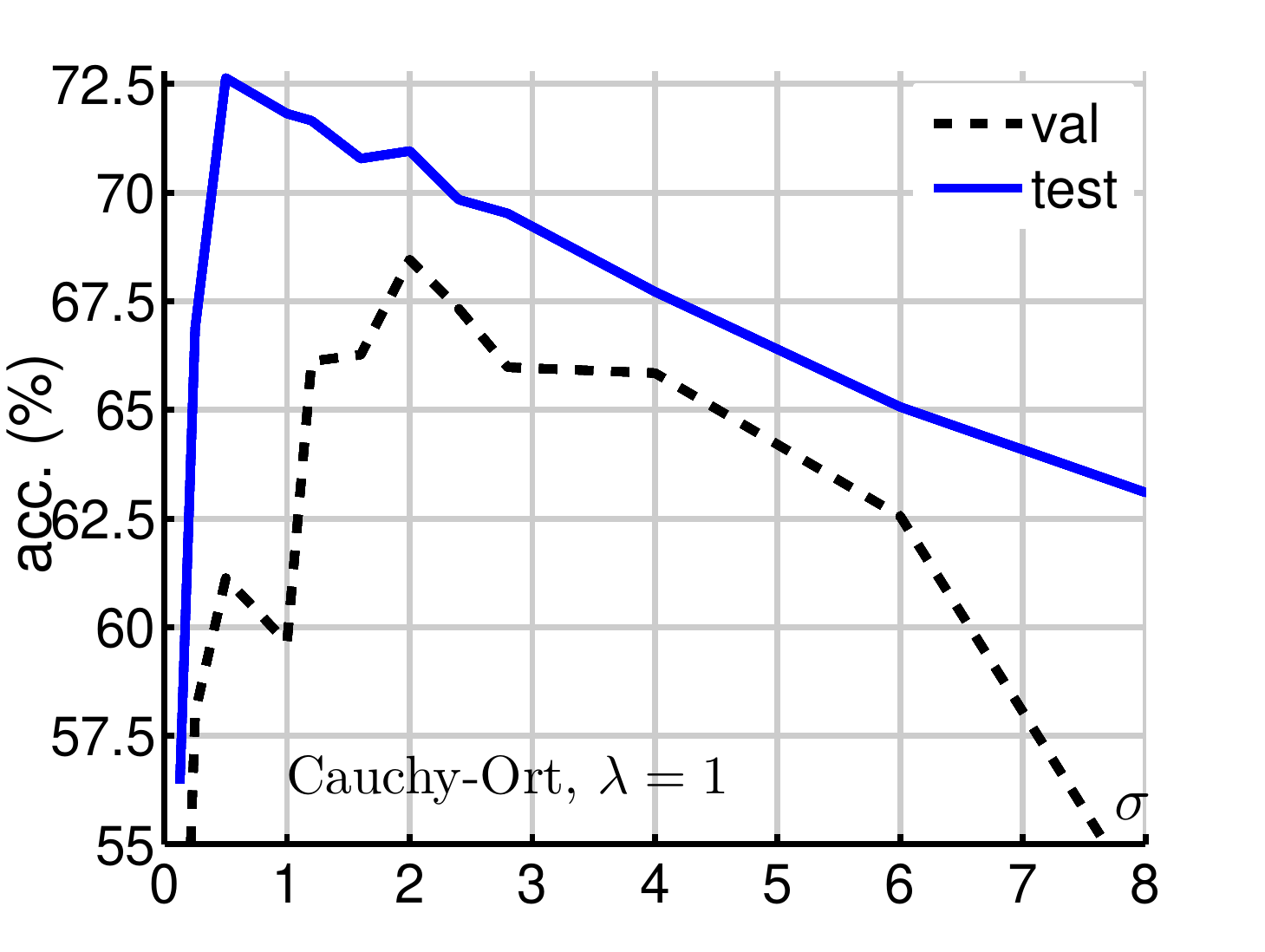}
\end{subfigure}
\begin{subfigure}[b]{\SensWW\linewidth}
\centering\includegraphics[trim=0 0 0 0, clip=true,height=\SensHHH]{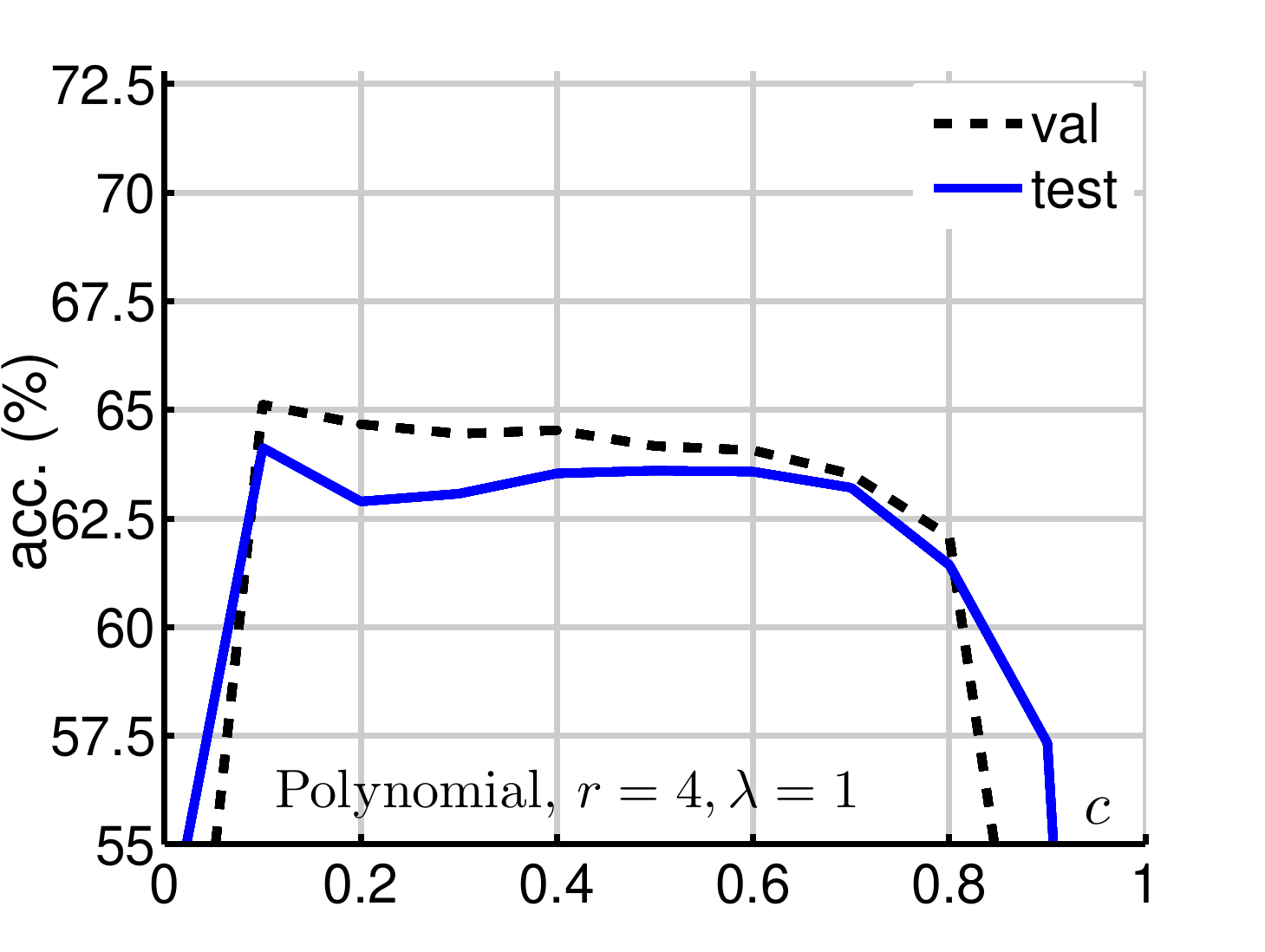}
\end{subfigure}
\hspace{0.1cm}
\begin{subfigure}[b]{\SensWW\linewidth}
\centering\includegraphics[trim=0 0 0 0, clip=true,height=\SensHHHH]{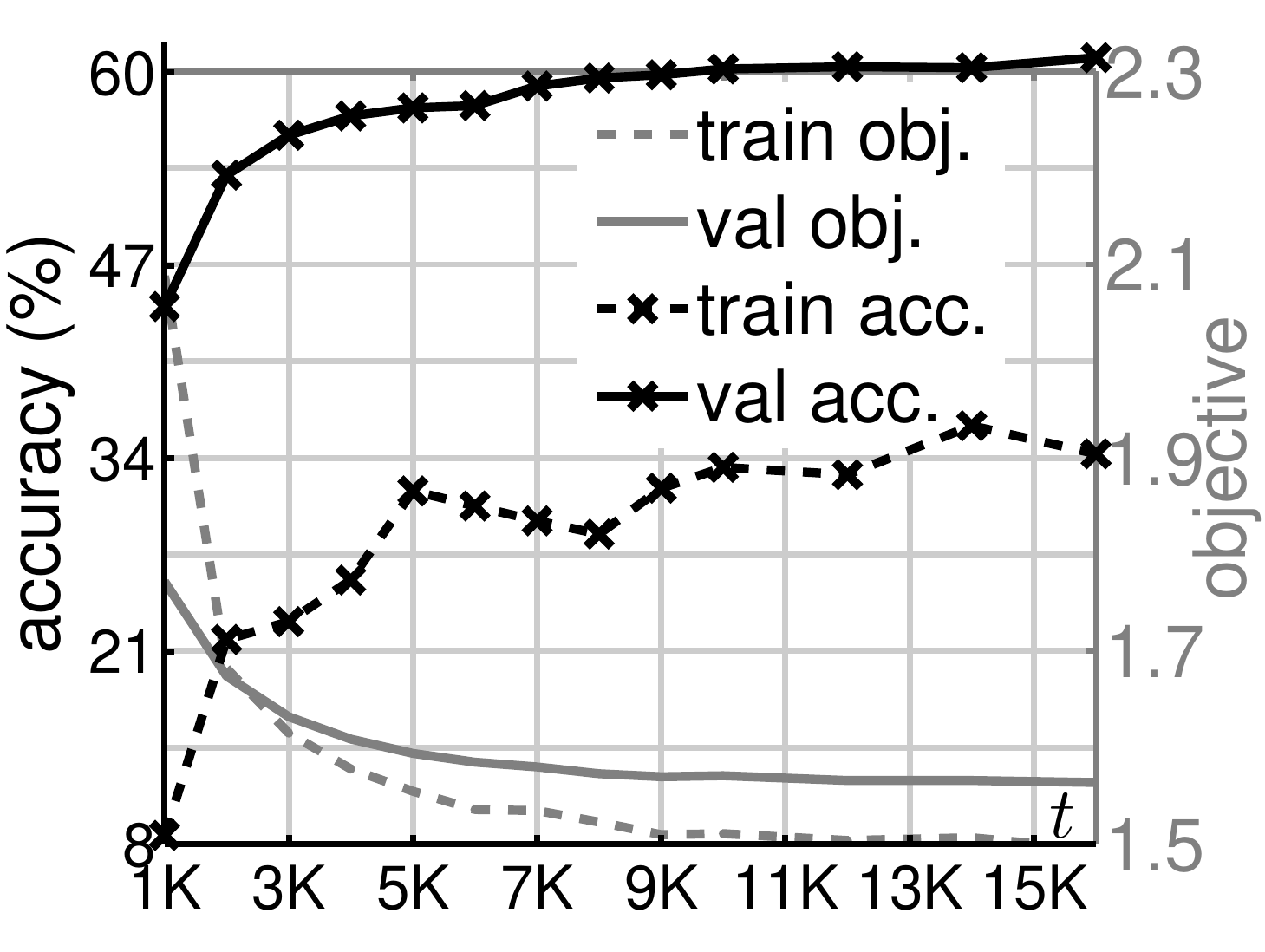}
\end{subfigure}
\\
\vspace{0.052cm}
\hspace{-0.3cm}
\begin{subfigure}[b]{\SensWW\linewidth}
\centering\includegraphics[trim=0 0 0 0, clip=true,height=\SensHHH]{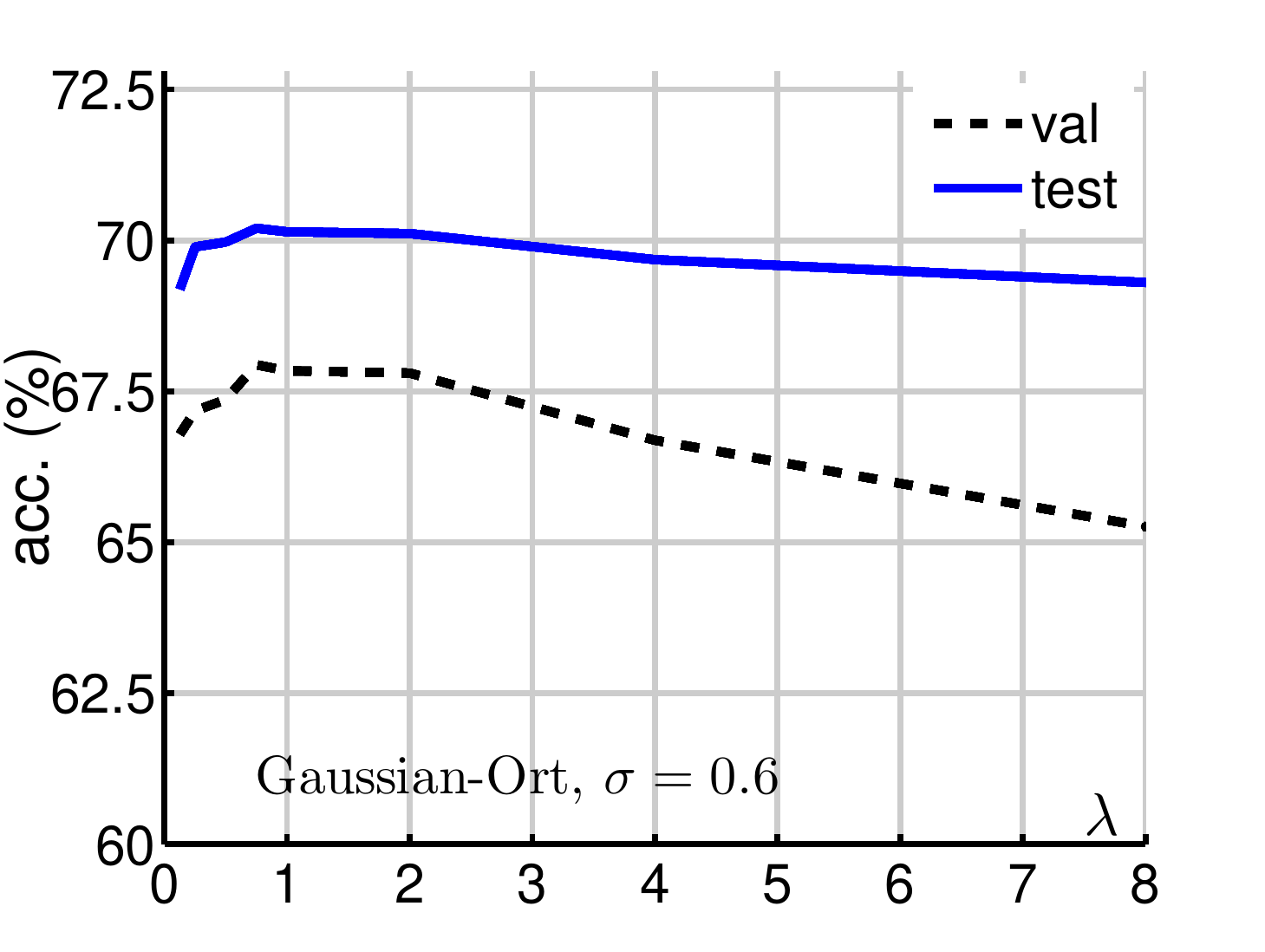}
\end{subfigure}
\begin{subfigure}[b]{\SensWW\linewidth}
\centering\includegraphics[trim=0 0 0 0, clip=true,height=\SensHHH]{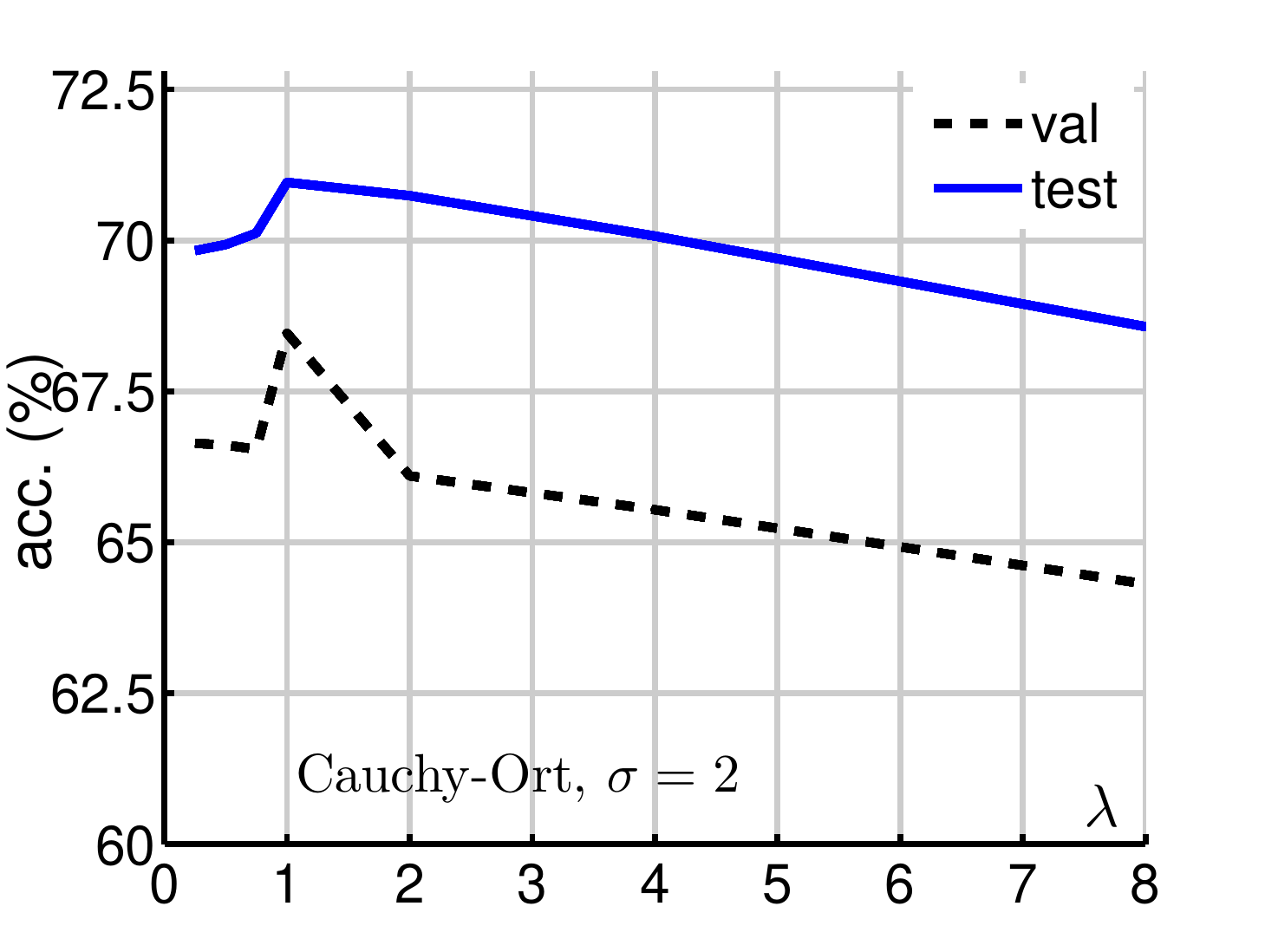}
\end{subfigure}
\begin{subfigure}[b]{\SensWW\linewidth}
\centering\includegraphics[trim=0 0 0 0, clip=true,height=\SensHHH]{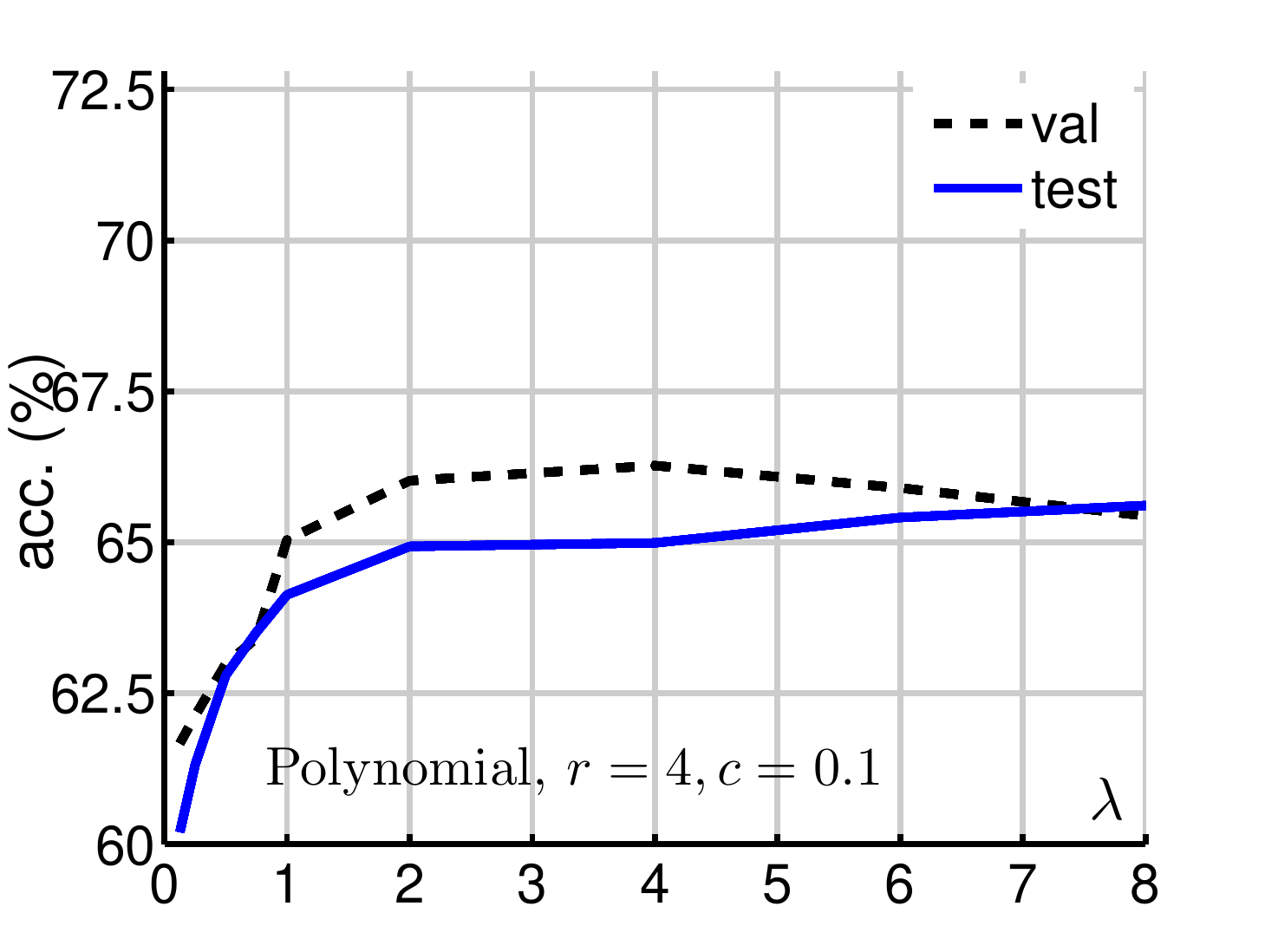}
\end{subfigure}
\hspace{0.1cm}
\begin{subfigure}[b]{\SensWW\linewidth}
\centering\includegraphics[trim=0 0 0 0, clip=true,height=\SensHHHH]{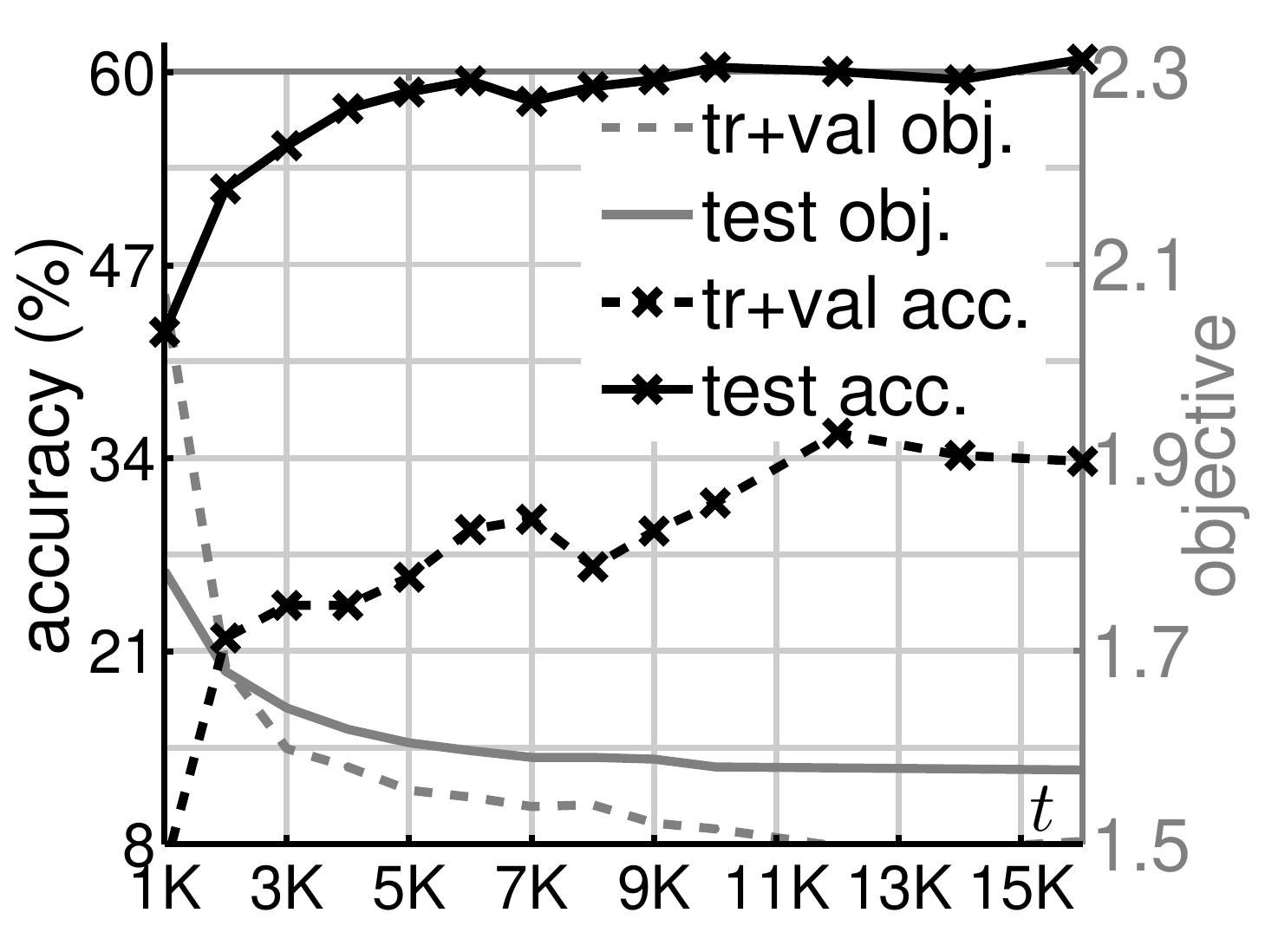}
\end{subfigure}
%
\caption{ (Columns 1--3) Validation ({\em val}) and testing ({\em test}) accuracies on AWA1 for the standard evaluation protocol. We 
vary $\sigma$, $\lambda$ and $c$. In the top row, we vary the radius $\sigma$ for kernels ({\em Gaussian-Ort}) and ({\em Cauchy-Ort}). For ({\em Polynomial}$, r\!=\!4$), we vary its bias $c$. The bottom row shows performance w.r.t. regularization $\lambda$ of our kernel polarization model given fixed $\sigma$ and/or $c$ selected earlier via cross-validation. (Column 4) The accuracy and the objective w.r.t. the number of iterations $t$ for ({\em Gaussian-Ort}) on the SUN dataset. The top and bottom plots concern training ({\em train}) vs. validation ({\em val}) and training plus validation ({\em tr+val}) vs. testing ({\em test}), respectively.
}\vspace{-0.3cm}
\label{fig:pars}
\end{figure*}

We note that our models which impose our soft/implicit incoherence outperform again the variants which do not impose it. Moreover, the Cauchy kernel appears to be an adequate choice for varied testing tasks. Cauchy may be less prone to overfitting to local clusters of datapoints as its tails decay slower compared to tails of Gaussian.

\vspace{0.05cm}
\noindent{\textbf{Sensitivity analysis.}} Robust algorithms are expected to generalize well to unseen data and avoid overfitting \eg, best classification scores on the validation and test data may be far away from each other when considered as a function of hyperparameters. Moreover, oversensitivity to the choice of hyperparameters may result in an algorithm which is hard to fine-tune.  Below, we show how our methods behave w.r.t. the choice of radius $\sigma$ for the shift-invariant Gaussian and Cauchy kernels, the bias parameter $c$ of the Polynomial kernel and the regularization parameter $\lambda$ from Eq. \eqref{eq:zero_polar}.

Figure \ref{fig:pars} (columns 1--3) shows how our zero-shot kernel learning performs on the AWA1 dataset w.r.t. the listed above hyperparameters. The standard evaluation protocol is used. The top row demonstrates that the kernel radius is an important parameter in our setup. For ({\em Gaussian-Ort}), the validation and test curves vary smoothly. The best results are attained for $\sigma\!=\!0.6$ and $\sigma\!=\!1$, respectively. The difference in the testing accuracy evaluated at $\sigma\!=\!0.6$ and $\sigma\!=\!1$ amounts to 2.5\%. The similar trend emerges for ({\em Cauchy-Ort}). Such a discrepancy between the validation and testing scores reflects that there is a visible domain shift between the validation and test problems. This result is typical for the knowledge transfer tasks such as domain adaptation and zero-shot learning. For ({\em Polynomial}, $r\!=\!4$), the validation and testing scores attain maximum for the same $c\!=\!1$, however, they are $\sim$5\% below performance of ({\em Gaussian-Ort}) and ({\em Cauchy-Ort}). This reveals that while the Polynomial kernel may generalize well, it lacks the capacity of the RBF kernels to capture highly non-linear data patterns.

Figure \ref{fig:pars} (columns 1--3, bottom) show our scores  w.r.t. the regularization parameter $\lambda$ which, in our setting, controls how strongly the between-class datapoints are pushed from each other after the projection into the attribute space. 

It appears that the peaks in validation and testing accuracies match each other well for all three kernels used in this experiment. The value of $\lambda$ does not affect dramatically the performance of our algorithm and the range for which the best performance is attained varies from 0.8--2. However, it is also clear that if $\lambda\!\rightarrow\!0$ or $\lambda\!\rightarrow\!\infty$, the scores drop. This demonstrates the importance of balancing the impact of so-called within- and between-class statistics in our zero-shot kernel model which `polarizes' these statistics.

Figure \ref{fig:pars} (column 4) shows how our algorithm behaves w.r.t. the number of solver iterations $t$ on the SUN dataset. 
The top row shows that the training objective attains lower values compared to the validation objective for $t\!\geq\!2000$. As the objective decreases, there is a clear  increase in accuracy for both the training and validation curves. The same behavior is observed in the bottom row which demonstrates that our algorithm is stable w.r.t. to $t$.

Finally, Table \ref{tab:incoh_total} shows that Proposition \ref{prop:inc1} and following from it Eq. \eqref{eq:zero_polar} promote weak incoherence.

\section{Conclusions}
\label{sec:conclude}

In this paper, we have proposed a novel approach to zero-shot learning by the use of kernels. Our model is inspired by the the Linear Discriminant Analysis \cite{fisher_fda,duda_patternclass} and kernel alignment methods \cite{kernel_alignment,ploarisation}. To the best of our knowledge, we are the first to show how to combine zero-shot learning with the Polynomial and the RBF family of kernels to obtain a non-linear compatibility function. Our model shows that a learned projection that embeds datapoints in the attribute space and from there, in the non-linear Hilbert space, is a robust tool for zero-shot learning. We learn an approximate subspace by encouraging in a soft/implicit manner the incoherence between column vectors of the projection matrix. Therefore, our projection incorporates rotation and scaling but prevents shear which causes overfitting due to more degrees of freedom in such an unconstrained model.  

Each of our models achieve state-of-the-art results on up to four out of five datasets on the standard zero-shot learning benchmark for new stricter recently proposed datasplits. Moreover, each of our models obtain state-of-the-art results on up to three out of five datasets on the new generalized zero-shot learning benchmark which takes into account so-called harmonized scores for classes seen and unseen during the training process. 
We note that if we were to pick one best kernel per dataset, this would lead to further improvements in accuracy. For future directions, this warrants an investigation into multiple kernel learning in the context of zero-shot kernel learning. We also plan to investigate the benefit of learning class-wise subspace matrices.

\ifdefined\arxiv
\begin{table}[t]
\else
\begin{table}[b]
\fi
\centering
%
{
\renewcommand{\arraystretch}{0.8}
{
\fontsize{9}{10}\selectfont
\centering
\hspace{-0.3cm}
\begin{tabular}{c | c c c c c } 
\hline
Gauss.     & AWA1  & AWA2  & SUN   & CUB & aPY   \\
{\em Non-Ort} & 376.4 & 420.7 & 209.1 & 2361.4 & 342.1 \\
{\em Eq. \eqref{eq:zero_polar}}    & {\bf 131.6} & {\bf 205.0} & {\bf 59.1} & {\bf 1018.8} & {\bf 259.5} \\
\hline
Cauchy & AWA1  & AWA2  & SUN  & CUB & aPY  \\
{\em Non-Ort} & 357.7 & 346.9 & 187.1 & 2614.4 & 334.0\\
{\em Eq. \eqref{eq:zero_polar}}  & {\bf 138.3} & {\bf 178.2} & {\bf 124.1} & {\bf 1227.5} & {\bf 214.9}\\
\hline
\end{tabular}
}
}
\caption{Incoherence on various kernels and datasets. For the $\ell_2$-norm normalized columns of $\mW$ we computed $||\mW^T\!\mW\!-\!\mIdent||_F^2$. Lower values indicate more incoherence between columns of $\mW$.}
\label{tab:incoh_total}
\vspace{-0.3cm}
\end{table}

\renewcommand*\appendixpagename{Appendix}
\begin{appendices}

\ifdefined\arxiv
\renewcommand{\SrcImgWW}{0.104}
\renewcommand{\SrcImgWWW}{0.058}
\newcommand{\SrcImgHH}{1.36cm}
\else
	\ifdefined\SrcImgWW
	\renewcommand{\SrcImgWW}{0.104}
	\else
	\newcommand{\SrcImgWW}{0.104}
	\fi

	\ifdefined\SrcImgWWW
	\renewcommand{\SrcImgWWW}{0.058}
	\else
	\newcommand{\SrcImgWWW}{0.058}
	\fi

	\ifdefined\SrcImgHH
	\renewcommand{\SrcImgHH}{1.36cm}
	\else
	\newcommand{\SrcImgHH}{1.36cm}
	\fi

\fi

\newcommand{\RE}{\color{blue!20!black!30!red}}

\newcommand{\SSS}{{\em S}}
\newcommand{\TTT}{{\em T}}
\newcommand{\SPT}{{\footnotesize\em S+T}}
\newcommand{\SOO}{{\em So}}
\newcommand{\JBL}{{\footnotesize\em JBLD}}
\newcommand{\JBD}{{\em JBLD}}
\newcommand{\AIR}{{\footnotesize\em AIRM}}

\newcommand{\BL}{\color{black!40!green}}
\newcommand{\BU}[1]{\color{black!40!green}\textbf{#1}}
\newcommand{\BW}[1]{\color{black!40!green}#1}
\newcommand{\BO}[1]{\textbf{#1}}
\newcommand{\TKN}[2]{{\footnotesize top-{#1}-{#2}}}
\newcommand{\TKO}{\footnotesize\pbox{3cm}{$\avg_k$\\top-$k$-$k$}}

\newcommand{\BV}[1]{\color{black!10!blue}{\em #1}}

\section{Weak incoherence of $\mW$}
\label{sec:ort}

\newcommand{\absval}[1]{\ifnum#1<0 -\fi#1}
\newcommand{\signval}[1]{\ifnum#1<0 -\fi1}

\newcommand\ctr[2]{\pgfmathsetmacro{\compB}{0.5+(#1-#2)/(#1>#2?#1:#2)}\color{black}\edef\x{\noexpand\cellcolor[hsb]{0,\compB,1}}\x#1}

\newcommand\ctb[2]{\pgfmathsetmacro{\compB}{#1/#2}\color{black}\edef\x{\noexpand\cellcolor[hsb]{0.95,\compB,1}}\x#1}

\newcommand\ctc[2]{\pgfmathsetmacro{\compB}{#1/#2}\color{black}\edef\x{\noexpand\cellcolor[hsb]{0.7,\compB,1}}\x#1}

\newcommand\ctd[2]{\pgfmathsetmacro{\compB}{\ifnum#1>#2 0.5+(#1-#2)/#1/2\else 0.5+(#1-#2)/#2/2\fi}\color{black}\edef\x{\noexpand\cellcolor[hsb]{0.55,\compB,1}}\x#1}

\newcommand\ctbb[2]{\pgfmathsetmacro{\compB}{(#1>0?#1:-#1)/#2}\pgfmathsetmacro{\compC}{#1/1}\color{black}\edef\x{\noexpand\cellcolor[hsb]{0,\compB,1}}\x\compC}
\newcommand\ctbg[2]{\pgfmathsetmacro{\compB}{(#1>0?#1:-#1)/#2}\pgfmathsetmacro{\compC}{#1/1}\color{black}\edef\x{\noexpand\cellcolor[hsb]{0,\compB,1}}\x\compC}

\begin{table}[h]
\setlength{\tabcolsep}{0.1em}
\fontsize{7}{9}\selectfont
\centering
\begin{tabular}{|c c c c c c c c c c|}
\multicolumn{10}{|c|}{\cellcolor[rgb]{0.9,0.9,0.9}{\em Gaussian-Ort}}\\
   \ctbb{1.0000}{1}  & \ctbb{0.0384}{1}   &   \ctbb{-0.0368}{1}  &  \ctbb{-0.0940}{1}  &   \ctbb{-0.1192}{1}  &   \ctbb{-0.2234}{1}  &   \ctbb{-0.0582}{1}  &   \ctbb{-0.1484}{1}  &  \ctbb{0.0876}{1}   &  \ctbb{0.0628}{1}   \\
   \ctbb{0.0384}{1}  & \ctbb{1.0000}{1}   &   \ctbb{0.0075}{1}   &  \ctbb{-0.2262}{1}  &   \ctbb{-0.0523}{1}  &   \ctbb{-0.0172}{1}  &   \ctbb{-0.0501}{1}  &   \ctbb{-0.0689}{1}  &  \ctbb{0.0674}{1}   &  \ctbb{0.1292}{1}   \\
   \ctbb{-0.0368}{1} & \ctbb{0.0075}{1}   &   \ctbb{1.0000}{1}   &  \ctbb{-0.0715}{1}  &   \ctbb{0.1992}{1}   &   \ctbb{0.0166}{1}   &   \ctbb{0.0715}{1}   &   \ctbb{0.0342}{1}   &  \ctbb{-0.1668}{1}  &  \ctbb{0.0438}{1}   \\
   \ctbb{-0.0940}{1} & \ctbb{-0.2262}{1}  &   \ctbb{-0.0715}{1}  &  \ctbb{1.0000}{1}   &   \ctbb{0.0912}{1}   &   \ctbb{-0.0582}{1}  &   \ctbb{-0.0530}{1}  &   \ctbb{0.0621}{1}   &  \ctbb{-0.0905}{1}  &  \ctbb{-0.1422}{1}  \\
   \ctbb{-0.1192}{1} & \ctbb{-0.0523}{1}  &   \ctbb{0.1992}{1}   &  \ctbb{0.0912}{1}   &   \ctbb{1.0000}{1}   &   \ctbb{-0.1231}{1}  &   \ctbb{-0.0914}{1}  &   \ctbb{-0.1321}{1}  &  \ctbb{0.0249}{1}   &  \ctbb{-0.0650}{1}  \\
   \ctbb{-0.2234}{1} & \ctbb{-0.0172}{1}  &   \ctbb{0.0166}{1}   &  \ctbb{-0.0582}{1}  &   \ctbb{-0.1231}{1}  &   \ctbb{1.0000}{1}   &   \ctbb{0.0539}{1}   &   \ctbb{-0.0298}{1}  &  \ctbb{-0.0015}{1}  &  \ctbb{0.0740}{1}   \\
   \ctbb{-0.0582}{1} & \ctbb{-0.0501}{1}  &   \ctbb{0.0715}{1}   &  \ctbb{-0.0530}{1}  &   \ctbb{-0.0914}{1}  &   \ctbb{0.0539}{1}   &   \ctbb{1.0000}{1}   &   \ctbb{-0.1084}{1}  &  \ctbb{-0.0629}{1}  &  \ctbb{0.0396}{1}   \\
   \ctbb{-0.1484}{1} & \ctbb{-0.0689}{1}  &   \ctbb{0.0342}{1}   &  \ctbb{0.0621}{1}   &   \ctbb{-0.1321}{1}  &   \ctbb{-0.0298}{1}  &   \ctbb{-0.1084}{1}  &   \ctbb{1.0000}{1}   &  \ctbb{0.0321}{1}   &  \ctbb{0.3139}{1}   \\
   \ctbb{0.0876}{1}  & \ctbb{0.0674}{1}   &   \ctbb{-0.1668}{1}  &  \ctbb{-0.0905}{1}  &   \ctbb{0.0249}{1}   &   \ctbb{-0.0015}{1}  &   \ctbb{-0.0629}{1}  &   \ctbb{0.0321}{1}   &  \ctbb{1.0000}{1}   &  \ctbb{0.0290}{1}   \\
   \ctbb{0.0628}{1}  & \ctbb{0.1292}{1}   &   \ctbb{0.0438}{1}   &  \ctbb{-0.1422}{1}  &   \ctbb{-0.0650}{1}  &   \ctbb{0.0740}{1}   &   \ctbb{0.0396}{1}   &   \ctbb{0.3139}{1}   &  \ctbb{0.0290}{1}   &  \ctbb{1.0000}{1}   \\
\hline
\end{tabular}
\caption{Illustration of the first $10\!\times\!10$ elements of $\mW^T\!\mW$ for ({\em Gaussian-Ort}). We $\ell_2$-norm normalized columns of $\mW$ and then color-coded cells. The intense red indicates closeness of the off-diagonal values to one. As can be seen, the off-diagonal entries have much lower values than the elements on the diagonal.}\label{tab:ort_case}
\end{table}

\begin{table}[h]
\setlength{\tabcolsep}{0.1em}
\fontsize{7}{9}\selectfont
\centering
\begin{tabular}{|c c c c c c c c c c|}
\multicolumn{10}{|c|}{\cellcolor[rgb]{0.9,0.9,0.9}{\em Gaussian}}\\
   \ctbg{1.0000}{1} &    \ctbg{0.2590}{1} &   \ctbg{-0.0556}{1} &   \ctbg{-0.1720}{1} &   \ctbg{-0.1865}{1} &   \ctbg{-0.1168}{1} &   \ctbg{-0.1503}{1} &   \ctbg{-0.1406}{1} &    \ctbg{0.2686}{1} &    \ctbg{0.2076}{1}\\
    \ctbg{0.2590}{1} &    \ctbg{1.0000}{1} &    \ctbg{0.0257}{1} &   \ctbg{-0.4768}{1} &   \ctbg{-0.1866}{1} &   \ctbg{-0.0960}{1} &   \ctbg{-0.1888}{1} &   \ctbg{-0.0923}{1} &    \ctbg{0.3299}{1} &    \ctbg{0.2592}{1}\\
   \ctbg{-0.0556}{1} &    \ctbg{0.0257}{1} &    \ctbg{1.0000}{1} &   \ctbg{-0.3542}{1} &    \ctbg{0.2656}{1} &   \ctbg{-0.1339}{1} &   \ctbg{-0.0388}{1} &   \ctbg{-0.0986}{1} &   \ctbg{-0.1592}{1} &   \ctbg{-0.1455}{1}\\
   \ctbg{-0.1720}{1} &   \ctbg{-0.4768}{1} &   \ctbg{-0.3542}{1} &    \ctbg{1.0000}{1} &   \ctbg{-0.1488}{1} &   \ctbg{-0.0246}{1} &    \ctbg{0.0480}{1} &   \ctbg{-0.0241}{1} &    \ctbg{0.0021}{1} &   \ctbg{-0.0497}{1}\\
   \ctbg{-0.1865}{1} &   \ctbg{-0.1866}{1} &    \ctbg{0.2656}{1} &   \ctbg{-0.1488}{1} &    \ctbg{1.0000}{1} &   \ctbg{-0.2122}{1} &   \ctbg{-0.0162}{1} &   \ctbg{-0.2212}{1} &   \ctbg{-0.1753}{1} &   \ctbg{-0.2876}{1}\\
   \ctbg{-0.1168}{1} &   \ctbg{-0.0960}{1} &   \ctbg{-0.1339}{1} &   \ctbg{-0.0246}{1} &   \ctbg{-0.2122}{1} &    \ctbg{1.0000}{1} &    \ctbg{0.6296}{1} &    \ctbg{0.4083}{1} &   \ctbg{-0.0570}{1} &   \ctbg{-0.0863}{1}\\
   \ctbg{-0.1503}{1} &   \ctbg{-0.1888}{1} &   \ctbg{-0.0388}{1} &    \ctbg{0.0480}{1} &   \ctbg{-0.0162}{1} &    \ctbg{0.6296}{1} &    \ctbg{1.0000}{1} &   \ctbg{-0.0041}{1} &   \ctbg{-0.0791}{1} &   \ctbg{-0.0859}{1}\\
   \ctbg{-0.1406}{1} &   \ctbg{-0.0923}{1} &   \ctbg{-0.0986}{1} &   \ctbg{-0.0241}{1} &   \ctbg{-0.2212}{1} &    \ctbg{0.4083}{1} &   \ctbg{-0.0041}{1} &    \ctbg{1.0000}{1} &    \ctbg{0.0188}{1} &    \ctbg{0.1778}{1}\\
    \ctbg{0.2686}{1} &    \ctbg{0.3299}{1} &   \ctbg{-0.1592}{1} &    \ctbg{0.0021}{1} &   \ctbg{-0.1753}{1} &   \ctbg{-0.0570}{1} &   \ctbg{-0.0791}{1} &    \ctbg{0.0188}{1} &    \ctbg{1.0000}{1} &    \ctbg{0.3737}{1}\\
    \ctbg{0.2076}{1} &    \ctbg{0.2592}{1} &   \ctbg{-0.1455}{1} &   \ctbg{-0.0497}{1} &   \ctbg{-0.2876}{1} &   \ctbg{-0.0863}{1} &   \ctbg{-0.0859}{1} &    \ctbg{0.1778}{1} &    \ctbg{0.3737}{1} &    \ctbg{1.0000}{1}\\
\hline
\end{tabular}
\caption{Illustration of the first $10\!\times\!10$ elements of $\mW^T\!\mW$ for ({\em Gaussian}). We $\ell_2$-norm normalized columns of $\mW$ and then color-coded cells. The intense red indicates closeness of the off-diagonal values to one. As can be seen, the off-diagonal entries have values that are sometimes comparable to the values on the diagonal.}\label{tab:nonort_case}
\end{table}

To demonstrate that our solution learns $\mW$ with weakly incoherent w.r.t. each other column vectors $\vw_1,\cdots,\vw_{d'}\!$ for ({\em Gaussian-Ort}), we display the first $10\!\times\!10$ entries of $\mW^T\!\mW$ for $\mW$ with $\ell_2$-norm normalized column vectors. We learned $\mW$ on the AWA1 dataset. As can be seen from Table \ref{tab:ort_case}, the values of the off-diagonal elements are much lower than the values of the elements on the diagonal. This result implies the weak incoherence between the column vectors of $\mW$. 
In contrast, Table \ref{tab:nonort_case} presents the first $10\!\times\!10$ entries of $\mW^T\!\mW$, where $\mW$ contains $\ell_2$-norm normalized column vectors, for ({\em Gaussian}) which does not impose the soft/implicit incoherence constraints. We display results for $\mW$ learned on the AWA1 dataset. 
As can be seen from the table, the values of the off-diagonal elements are often comparable to the values of the elements on the diagonal. This result implies there is no incoherence between the column vectors of $\mW$.

\end{appendices}


{\small

}

\end{document}